\documentclass[10pt,journal,compsoc]{IEEEtran}

\ifCLASSOPTIONcompsoc
  \usepackage[nocompress]{cite}
\else
  \usepackage{cite}
\fi
\ifCLASSINFOpdf
\else
\fi
\usepackage{amsmath,amsfonts}
\usepackage{algorithmic}
\usepackage{algorithm}
\usepackage{array}
\usepackage[caption=false,font=normalsize,labelfont=sf,textfont=sf]{subfig}
\usepackage{textcomp}
\usepackage{stfloats}
\usepackage{url}
\usepackage{verbatim}
\usepackage{graphicx}
\usepackage{cite}
\usepackage{xfp,graphicx}
\usepackage{tabularx}
\usepackage{bm}
\usepackage[pagebackref=true,breaklinks=true,colorlinks,linkcolor=blue,citecolor=blue,bookmarks=false]{hyperref}
\usepackage{wrapfig}
\usepackage{mathtools}
\usepackage{annotate-equations}
\usepackage{custom}
\usepackage{lipsum}
\usepackage{pifont}
\usepackage[misc]{ifsym}

\begin{document}

\title{DreamWaltz-G: Expressive 3D Gaussian Avatars from Skeleton-Guided 2D Diffusion}

\author{
    Yukun~Huang,
    Jianan~Wang,
    Ailing~Zeng,~\IEEEmembership{Member,~IEEE,}
    Zheng-Jun~Zha,~\IEEEmembership{Member,~IEEE,}
    Lei~Zhang,~\IEEEmembership{Fellow,~IEEE,}
    Xihui~Liu\textsuperscript{\Letter}~\IEEEmembership{Member,~IEEE}
    \IEEEcompsocitemizethanks{
        \IEEEcompsocthanksitem Y. Huang and X. Liu are with The University of Hong Kong (HKU), Hong Kong SAR 999077, China.\\
        E-mail: yukun@hku.hk, xihuiliu@eee.hku.hk
        \IEEEcompsocthanksitem J. Wang is with Astribot, Shenzhen 518063, China.\\
        E-mail: jiananwang@astribot.com
        \IEEEcompsocthanksitem A. Zeng is with Tencent, Shenzhen 518054, China.\\
        E-mail: ailingzengzzz@gmail.com
        \IEEEcompsocthanksitem Z. Zha is with University of Science and Technology of China (USTC), Hefei 230026, China.\\
        E-mail: zhazj@ustc.edu.cn
        \IEEEcompsocthanksitem L. Zhang is with International Digital Economy Academy (IDEA), Shenzhen 518045, China.\\
        E-mail: leizhang@idea.edu.cn
    }
    \thanks{\Letter~: Corresponding author.}
}

\IEEEtitleabstractindextext{
\begin{abstract}
Leveraging pretrained 2D diffusion models and score distillation sampling (SDS), recent methods have shown promising results for text-to-3D avatar generation. However, generating high-quality 3D avatars capable of expressive animation remains challenging. In this work, we present DreamWaltz-G, a novel learning framework for animatable 3D avatar generation from text. The core of this framework lies in Skeleton-guided Score Distillation and Hybrid 3D Gaussian Avatar representation. Specifically, the proposed skeleton-guided score distillation integrates skeleton controls from 3D human templates into 2D diffusion models, enhancing the consistency of SDS supervision in terms of view and human pose. This facilitates the generation of high-quality avatars, mitigating issues such as multiple faces, extra limbs, and blurring. The proposed hybrid 3D Gaussian avatar representation builds on the efficient 3D Gaussians, combining neural implicit fields and parameterized 3D meshes to enable real-time rendering, stable SDS optimization, and expressive animation. Extensive experiments demonstrate that DreamWaltz-G is highly effective in generating and animating 3D avatars, outperforming existing methods in both visual quality and animation expressiveness. Our framework further supports diverse applications, including human video reenactment and multi-subject scene composition. For more vivid 3D avatar and animation results, please visit \href{https://yukun-huang.github.io/DreamWaltz-G/}{https://yukun-huang.github.io/DreamWaltz-G/}.
\end{abstract}

}

\maketitle

\IEEEdisplaynontitleabstractindextext
\IEEEpeerreviewmaketitle


\IEEEraisesectionheading{\section{Introduction}\label{sec:intro}}

\begin{figure*}[htbp]
\centering
\vspace{-2em}
\includegraphics[width=1.\linewidth]{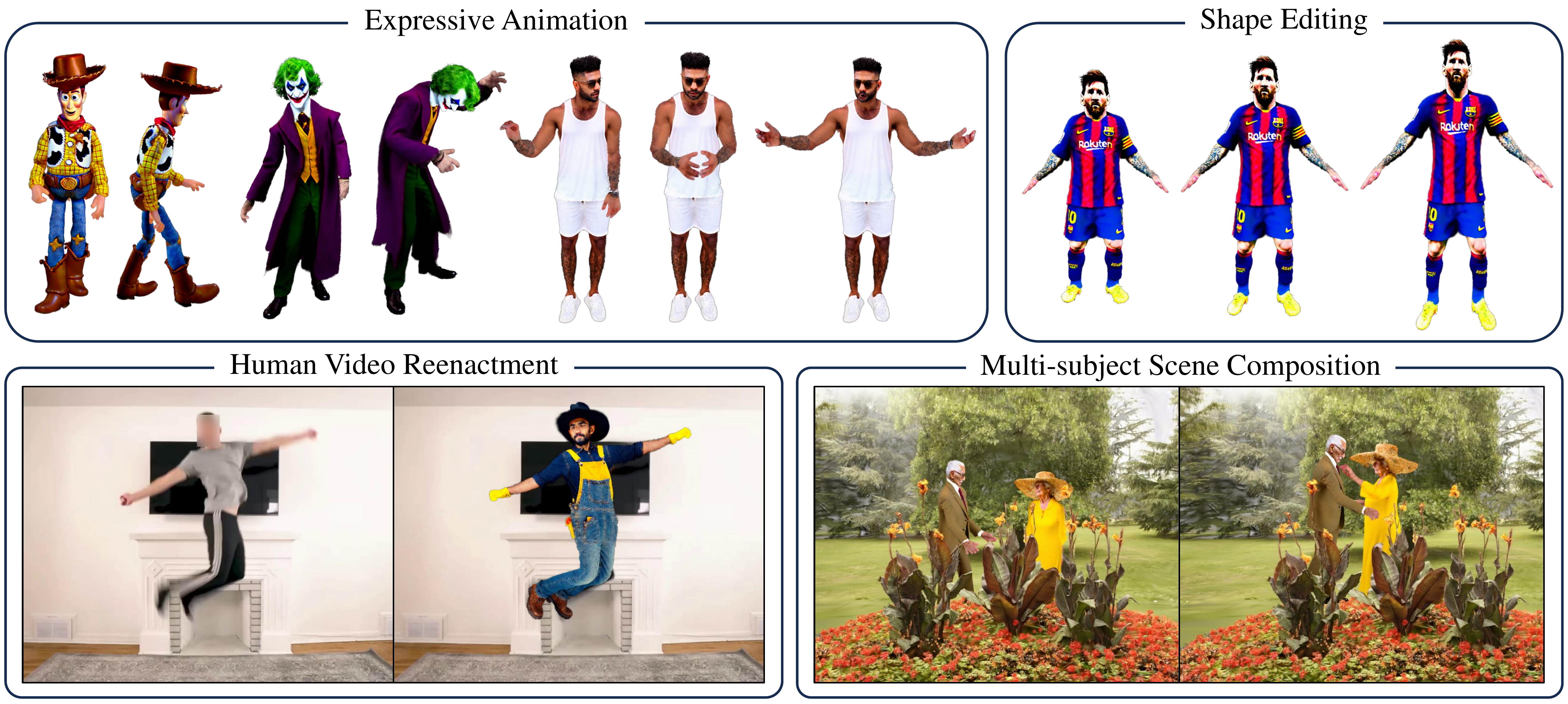}
\caption{We present DreamWaltz-G, a text-driven animatable 3D avatar generation framework, which can create high-quality 3D avatars from imaginative text prompts and animate them given motion sequences without manual rigging and retraining. Our method enables various downstream applications, such as expressive animation production, shape editing, human video reenactment, and multi-subject scene composition.}
\label{fig:teaser}
\end{figure*}

\IEEEPARstart{A}{nimatable} 3D avatar generation is essential for a wide range of applications, such as film and cartoon production, video game design, and immersive media such as virtual/augmented reality. Traditional techniques for creating such intricate 3D avatars are costly and time-consuming, requiring thousands of hours from skilled artists with extensive aesthetics and 3D modeling knowledge.
Meanwhile, the advancement of 3D reconstruction~\cite{nerf,neus,dmtet,3dgs} has enabled promising methods which can reconstruct 3D human models from monocular images~\cite{saito2019pifu,xiu2022icon,xiu2023econ,weng2023personnerf,wang2023complete}, monocular videos~\cite{weng2022humannerf,jiang2022neuman,yu2023monohuman,qian20243dgsavatar}, or 3D scans~\cite{d3ga,zhao2022human,jiang2024uvgaussian,zheng2024physavatar}.
Nonetheless, these methods rely heavily on the collection of image/video data captured with a monocular camera or a synchronized camera array. This makes them unsuitable for generating 3D avatars from imaginative but abstract prompts like texts.

Recently, integrating pretrained text-to-image diffusion models~\cite{dalle2,latentdiffusion} into 3D modeling with score distillation sampling (SDS)~\cite{dreamfusion,sjc} has gained significant attention to make 3D digitization more accessible, alleviating the need for data collection. However, creating 3D avatars using a 2D diffusion model remains challenging. First, \textit{static avatars} require articulated structures with intricate parts (e.g., hands and faces) and detailed textures, which pretrained diffusion models and score distillation struggle to generate. Secondly, \textit{dynamic avatars} assume various poses in a coordinated and constrained manner, where changes in shape and appearance should be realistic without artifacts caused by inaccurate skeleton rigging. Although previous methods~\cite{avatarclip,jiang2023avatarcraft,liao2024tada,kolotouros2024dreamhuman,yuan2024gavatar,liu2024humangaussian,huang2023dreamwaltz} have demonstrated impressive results on text-driven 3D avatar creation, they still struggle with producing intricate geometric structures and detailed appearances, let alone for realistic animation.

In this paper, we present \textbf{DreamWaltz-G}, a zero-shot learning framework for text-driven 3D avatar generation. At the core of this framework are \textbf{Skel}eton-guided \textbf{S}core \textbf{D}istillation ({SkelSD}) and \textbf{H}ybrid \textbf{3}D \textbf{G}aussian~\cite{3dgs} \textbf{A}vatars ({H3GA}) for stable optimization and expressive animation.

For \textbf{SkelSD}, different from previous methods~\cite{liao2024tada,kolotouros2024dreamhuman,yuan2024gavatar} that only apply human priors to 3D avatar representations (\eg, 3D mesh~\cite{liao2024tada}), we additionally inject human priors into diffusion model through skeleton control~\cite{controlnet,humansd}, leading to a more stable SDS that conforms to the 3D human body structure. This design brings three benefits: (1) skeleton guidance from 3D human templates~\cite{smpl,smplx} enhances the 3D consistency of SDS and prevents the Janus (multi-face) problem; (2) it eliminates pose uncertainty of SDS and avoids defects such as extra limbs and ghosting; (3) randomly posed skeleton guidance enables pose-dependent shape and appearance learning from 2D diffusion model.

\textbf{H3GA} is a hybrid 3D representation for animatable 3D avatars, specifically designed to adapt SDS optimization and enable expressive animation. Specifically, H3GA combines the efficiency of 3D Gaussian Splatting~\cite{3dgs}, the local continuity of neural implicit fields~\cite{nerf,neus}, and the geometric accuracy of parameterized meshes~\cite{smpl,smplx}. As a result, H3GA supports real-time rendering, is robust to SDS optimization, and enables expressive animation with finger movements and facial expressions. Furthermore, considering the dynamic characteristics of different body parts, we designed a dual-branch deformation strategy to drive canonical 3D Gaussians for realistic animation.

Based on the proposed SkelSD and H3GA, DreamWaltz-G generates animatable 3D avatars in two training stages:

\noindent \textbf{(I) Canonical Avatar Generation}. For Stage I, we aim to create a canonical 3D avatar given text descriptions. Specifically, we employ Instant-NGP~\cite{instant-ngp} as the canonical avatar representation and optimize it with SkelSD for shape and appearance learning, where the skeleton guidance is extracted from SMPL-X~\cite{smplx} in the canonical pose.

\noindent \textbf{(II) Animatable Avatar Learning.} For Stage II, we aim to make the canonical avatar from Stage I rigged to SMPL-X and accurately animated. We employ H3GA as the animatable avatar representation for efficient deformation and stable optimization. Similar to Stage I, we use SkelSD for pose-dependent shape and appearance learning, except the skeleton guidance is extracted from SMPL-X in randomly sampled plausible poses.

In summary, our framework learns a hybrid 3D Gaussian avatar representation using skeleton-guided score distillation, ready for expressive animation and a wide range of applications, as illustrated in Figure~\ref{fig:teaser}. The key contributions of this work lie in four main aspects:
\begin{itemize}

\item We introduce a text-driven animatable 3D avatar generation framework, i.e., DreamWaltz-G, ready for expressive animation and various applications.

\item We propose SkelSD, a novel skeleton-guided score distillation strategy to reduce the view and pose inconsistencies between the 3D avatar's rendering and the 2D diffusion model's supervision.

\item We propose H3GA, a hybrid 3D Gaussian avatar representation that enables stable SDS optimization, real-time rendering, and expressive animation with finger movements and facial expressions.

\item Experiments demonstrate that DreamWaltz-G can effectively create animatable 3D avatars, achieving superior generation and animation quality compared to existing text-to-3D avatar methods.
\end{itemize}

Compared with the preliminary conference version~\cite{huang2023dreamwaltz}, this work introduces several non-trivial improvements. The most significant enhancement is the redesign of 3D avatar representation. Specifically, DreamWaltz~\cite{huang2023dreamwaltz} uses Instant-NGP~\cite{instant-ngp} for modeling 3D avatars. However, when applied to dynamic avatars with deformation, high-resolution sampling combined with inverse LBS~\cite{smpl} becomes computationally expensive and impractical for training. To address this, DreamWaltz-G adopts a novel hybrid 3D Gaussian representation, benefiting from efficient deformation and rendering of 3DGS~\cite{3dgs} while remaining compatible with SDS optimization and SMPL-X parameters.
Additionally, we replace the used 3D human parametric model SMPL~\cite{smpl} with SMPL-X~\cite{smplx}, introduce local geometric constraints for NeRF training, and explore more potential applications.

\section{Related Work}

\begin{table*}[htbp]
    \centering
    \caption{Comparisons of different text-driven 3D avatar generation methods. To clarify, \textit{Shape Control} refers to specifying the avatar's shape during generation instead of the shape initialization\textsuperscript{\dag}, while \textit{Shape Editing} involves adjusting the avatar's shape after generation.}
    \renewcommand\arraystretch{1.5}
    \begin{tabular}{l|c|c|c|c|c|c}
        \hline
        \textbf{Methods} & \textbf{3D Model} & \textbf{Body Animation} & \textbf{Hand Animation} & \textbf{Face Animation} & \textbf{Shape Control} & \textbf{Shape Editing} \\
        \hline
        DreamHuman~\cite{kolotouros2024dreamhuman} & NeRF & \checkmark & \xmark & \xmark & \xmark & \xmark \\
        DreamWaltz~\cite{huang2023dreamwaltz} & NeRF & \checkmark & \xmark & \xmark & \checkmark & \xmark \\
        TADA\textsuperscript{\dag}~\cite{liao2024tada} & Mesh & \checkmark & \checkmark & \checkmark & \xmark & \checkmark \\
        HumanGaussian~\cite{liu2024humangaussian} & 3DGS & \checkmark & \xmark & \xmark & \checkmark & \checkmark \\
        GAvatar~\cite{yuan2024gavatar} & 3DGS & \checkmark & \xmark & \xmark & \xmark & \checkmark \\
        \hline
        DreamWaltz-G (Ours) & 3DGS & \checkmark & \checkmark & \checkmark & \checkmark & \checkmark \\
        \hline
    \end{tabular}
    \label{tab:summary}
\end{table*}

We first review the previous methods for 2D diffusion models and then discuss recent advances in text-driven 3D object and 3D avatar generation.

\subsection{Text-driven Image Generation}
Recently, there have been significant advancements in text-to-image models such as GLIDE~\cite{glide}, unCLIP~\cite{dalle2}, Imagen~\cite{imagen}, and Stable Diffusion~\cite{latentdiffusion}, which enable the generation of highly realistic and imaginative images based on text prompts. 
These generative capabilities have been made possible by advancements in modeling, such as diffusion models~\cite{beatsgan, ddim, improved_ddpm}, 
and the availability of large-scale web data containing billions of image-text pairs~\cite{laion5b, cc, cc12}. 
These datasets encompass a wide range of general objects, with significant variations in color, texture, and camera viewpoints, providing pre-trained models with a comprehensive understanding of general objects and enabling the synthesis of high-quality and diverse objects.
Furthermore, recent works~\cite{controlnet, composer, humansd, xiao2024ccm} have explored incorporating additional conditioning, such as depth maps and human skeleton poses, to generate images with more precise control. With more advanced network architectures~\cite{dit,sdxl,sd3} and larger, higher-quality datasets~\cite{hyperhuman,objaverse}, the capabilities of text-to-image generation models continue to improve.

\subsection{Text-driven 3D Object Generation}
Dream Fields~\cite{dreamfields} and CLIPmesh~\cite{clipmesh} were groundbreaking in their utilization of CLIP~\cite{clip} to optimize an underlying 3D representation, aligning its 2D renderings with user-specified text prompts without necessitating costly 3D training data. 
However, this approach tends to result in less realistic 3D models since CLIP only provides discriminative supervision for high-level semantics. In contrast, recent works have demonstrated remarkable text-to-3D generation results by employing powerful text-to-image diffusion models as a robust 2D prior for optimizing a differentiable 3D representation with Score Distillation Sampling (SDS)~\cite{dreamfusion,sjc,magic3d,fantasia3d,dreamgaussian}. Nonetheless, the high variation in SDS leads to blurriness, over-saturated colors, and 3D inconsistencies. Although a series of subsequent works~\cite{dreamtime,nfsd,csd,liang2023luciddreamer,hifa,prolificdreamer} have introduced fundamental improvements to SDS optimization, the results remain unsatisfactory when applied to generating animatable 3D avatars with intricate details.

\subsection{Text-driven 3D Avatar Generation} 
Different from everyday objects, 3D avatars have detailed textures and intricate geometric structures that can be driven for realistic animation. Avatar-CLIP~\cite{avatarclip} employs CLIP~\cite{clip} for shape sculpting and texture generation but tends to produce less realistic and oversimplified 3D avatars. Unlike CLIP-based methods, both AvatarCraft~\cite{jiang2023avatarcraft} and DreamAvatar~\cite{cao2023dreamavatar} leverage powerful text-to-image diffusion models to provide 2D image guidance, effectively improving the visual quality of generated avatars. DreamWaltz~\cite{huang2023dreamwaltz} and AvatarVerse~\cite{zhang2024avatarverse} further utilizes ControlNet~\cite{controlnet} and SMPL~\cite{smpl} to provide view/pose-consistent 2D human guidance such as skeleton and DensePose~\cite{densepose}. Considering the limited 3D awareness of 2D diffusion models, HumanNorm~\cite{huang2024humannorm} proposes the normal-adapted and depth-adapted diffusion models for accurate geometry generation. In addition, to enable animatable avatar learning, DreamHuman~\cite{kolotouros2024dreamhuman} employs implicit 3D human model imGHUM~\cite{imghum} as 3D avatar representation, which improves the dynamic visual quality of generated avatars. Recently, 3D Gaussian Splatting (3DGS)~\cite{3dgs} has emerged as an explicit 3D representation enabling real-time deformation~\cite{yang2024deformable} and rendering. Some works~\cite{yuan2024gavatar,liu2024humangaussian,d3ga,jiang2024uvgaussian,hu2024gaussianavatar,moon2024expressive} have explored using 3DGS to represent 3D avatars. HumanGaussian~\cite{liu2024humangaussian} proposes a Structure-Aware SDS, which guides the adaptive density control of 3DGS with intrinsic human structures. GAvatar~\cite{yuan2024gavatar} introduces a primitive-based 3DGS representation where 3D Gaussians are defined inside pose-driven primitives to facilitate animation.

To highlight our contributions, we summarize the key differences between our work and related works in Table~\ref{tab:summary}.

\section{Method}\label{sec:method}

We first review some preliminary knowledge in Sec.~\ref{sec:method_preliminary}, then present the proposed {\emph{Skeleton-guided Score Distillation}} in Sec.~\ref{sec:method_skeleton_guided_sds} and {\emph{Hybrid 3D Gaussian Avatar Representation}} in Sec.~\ref{sec:method_hybrid_3dgs_avatar}. Finally, we introduce the text-driven 3D avatar generation framework \emph{DreamWaltz-G} in Sec.~\ref{sec:method_dreamwaltzG}.

\subsection{Preliminary}\label{sec:method_preliminary}
Before delving into our proposed method, we first introduce some concepts that form the basis of our framework.

\textbf{3D Gaussian Splatting} (3DGS)~\cite{3dgs} represents a 3D scene through a set of 3D Gaussians $\mathcal{G} = \{G_i \mid i = 1, \ldots, N\}$. The geometry of each 3D Gaussian $G_i$ is parameterized by a position (mean) $\mathbf{p}_i \in \mathbb{R}^{3 \times 1}$ and covariance matrix $\mathbf{\Sigma}_i \in \mathbb{R}^{3 \times 3}$ defined in world space:
\begin{equation*}
    G_i(\mathbf{x}) = e^{-\frac{1}{2} (\mathbf{x} - \mathbf{p}_i)^T \mathbf{\Sigma}_i^{-1} (\mathbf{x} - \mathbf{p}_i)},
\end{equation*}
where $\mathbf{x}$ is a 3D point in world coordinates.
To maintain the position semi-definite property of $\mathbf{\Sigma_i}$, a decomposition is used: $\mathbf{\Sigma}_i = \mathbf{R}_i \mathbf{S}_i \mathbf{S}_i^T \mathbf{R}_i^T$, where the scaling matrix $\mathbf{S}$ and the rotation matrix $\mathbf{R}$ are parameterized by a 3D vector $\mathbf{s}$ and a quaternion $\mathbf{q}$ for gradient descent.

To render an image, the 3D Gaussians can be projected to 2D using: $\bm{\Sigma}' = \mathbf{J} \mathbf{W} \mathbf{\Sigma} \mathbf{W}^T \mathbf{J}^T$, where $\mathbf{W}$ is a viewing transformation from world to camera coordinates, and $\mathbf{J}$ denotes the Jacobian of the affine approximation of the projective transformation. We use $G'_i$ parameterized by $\bm{\Sigma}'$ to represent the 2D Gaussian projected from $G_i$. Finally, the color $\mathbf{c}$ of each pixel $\mathbf{x}$ is rendered by alpha blending according to the 3D Gaussians' depth order $1,\ldots,N$:
\begin{equation*}
\mathbf{c}(\mathbf{x}) = \sum_{i=1}^{N} \mathbf{c}_i \alpha_i G'_i (\mathbf{x}) \prod_{j=1}^{i-1} (1 - \alpha_j G'_j (\mathbf{x})),
\end{equation*}
where $\alpha_i \in [0,1]$ is the opacity of $G_i$.

\textbf{Neural Radiance Field} (NeRF)~\cite{nerf,instant-ngp} is commonly used as the differentiable 3D representation for text-driven 3D generation~\cite{dreamfusion,magic3d}, parameterized by a trainable MLP. For rendering, a batch of rays $\mathbf{r}(k)=\mathbf{o}+k\mathbf{d}$ are sampled based on the camera position $\mathbf{o}$ and direction $\mathbf{d}$ on a per-pixel basis. The MLP takes $\mathbf{r}(k)$ as input and predicts density $\tau$ and color $c$. The volume rendering integral is then approximated using numerical quadrature to yield the final color of the rendered pixel: 
\begin{align*}
\hat{C}_c(\mathbf{r}) = \sum_{i=1}^{N_c} \Omega_i \cdot (1-\exp(-\tau_i\delta_i)) c_i,
\end{align*}
where $N_c$ is the number of sampled points on a ray, $\Omega_{i}=\exp(-\sum_{j=1}^{i-1}\tau_{j}\delta_{j})$ is the accumulated transmittance, and $\delta_i$ is the distance between adjacent sample points.

\textbf{Diffusion models}~\cite{ddpm,improved_ddpm} which have been pre-trained on extensive image-text datasets~\cite{dalle2,imagen,stable-dreamfusion} provide a robust image prior for supervising text-to-3D generation. Diffusion models learn to estimate the denoising score $\nabla_\mathbf{x}\log p_{\text{data}} (\mathbf{x})$ by adding noise to clean data $\mathbf{x} \sim p(\mathbf{x})$ (forward process) and learning to reverse the added noise (backward process). Noising the data distribution to isotropic Gaussian is performed in $T$ timesteps, with a pre-defined noising schedule $\alpha_t \in (0,1)$ and $\bar{\alpha}_t \coloneqq {\prod^t_{s=1}\alpha_s}$, according to:
\begin{align*}
\mathbf{x}_{t} =\sqrt{\bar{\alpha}_t} \mathbf{x} + \sqrt{1-\bar{\alpha}_{t}}\boldsymbol{\epsilon}, \text{ where } \boldsymbol{\epsilon} \sim \mathcal{N}(\mathbf{0}, \mathbf{I}).
\end{align*}
In the training process, the diffusion models learn to estimate the noise by
\begin{align*}
\mathcal{L}_{t} = \mathbb{E}_{\mathbf{x}, \boldsymbol{\epsilon} \sim \mathcal{N}(\mathbf{0},\mathbf{I}) }\left[\left\|\boldsymbol{\epsilon}_{\phi}\left(\mathbf{x}_t, t\right) - \boldsymbol{\epsilon} \right\|^{2}_{2}\right].
\end{align*}
Once trained, one can estimate $\mathbf{x}$ from noisy input and the corresponding noise prediction.

\textbf{Score Distillation} (SDS)~\cite{dreamfusion, magic3d, latentnerf} is a technique introduced by DreamFusion~\cite{dreamfusion} and extensively employed to distill knowledge from a pre-trained diffusion model $\boldsymbol{\epsilon}_\phi$ into a differentiable 3D representation. For a NeRF model parameterized by $\boldsymbol{\theta}$, its rendering $\mathbf{x}$ can be obtained by $\mathbf{x} = g(\boldsymbol{\theta})$ where $g$ is a differentiable renderer. SDS calculates the gradients of NeRF parameters $\boldsymbol{\theta}$ by,
\begin{align}
\quad\nabla_{\boldsymbol{\theta}}\mathcal{L}_{\text{SDS}}(\phi,\mathbf{x})=\mathbb{E}_{t, \boldsymbol{\epsilon}}\bigg[w(t)
(\boldsymbol{\epsilon}_\phi(\mathbf{x}_t;y,t)-\boldsymbol{\epsilon})\dfrac{\partial \mathbf{x}_t}{\partial\mathbf{x}}\dfrac{\partial \mathbf{x}}{\partial\boldsymbol{\theta}}\bigg],
\label{eq:sds}
\end{align}
where $w(t)$ is a weighting function that depends on the timestep $t$ and $y$ denotes the given text prompt.

\textbf{SMPL-X}~\cite{smplx} is a unified parametric 3D human model that extends SMPL~\cite{smpl} with fully articulated hands and an expressive face, containing $N_\text{v}=10,475$ vertices and $N_\text{j}=54$ joints.
Benefiting from its efficient and expressive human motion representation ability, SMPL-X has been widely used in human motion-driven tasks~\cite{avatarclip,zeng2022deciwatch,mahmood2019amass}.
The input parameters for SMPL-X include a 3D body joint and global rotation $\xi \in \mathbb{R}^{(N_\text{j}+1) \times 3}$, a body shape $\beta\in \mathbb{R}^{300}$, and a 3D global translation $t\in \mathbb{R}^{3}$. 

Formally, a triangulated mesh $T_\text{cnl}(\beta, \xi) \in \mathbb{R}^{N_\text{v} \times 3}$ in canonical pose is constructed by combining the template shape $\bar{T}$, the shape-dependent deformations $B_{S}(\beta)$, and the pose-dependent deformations $B_{P}(\xi)$ as,
\begin{equation}
T_\text{cnl}(\beta, \xi) = \bar{T} + B_{S}(\beta) + B_{P}(\xi),
\label{eq:smplx_cnl}
\end{equation}
where $B_{P}(\xi)$ is used to relieve artifacts in Linear Blend Skinning (LBS)~\cite{mohr2003building}.
Then, the LBS function is employed to transform the canonical mesh $T_\text{cnl}(\beta, \xi)$ into a triangulated mesh $T_\text{obs}(\beta, \xi)$ in the observed pose as,
\begin{equation}
T_\text{obs}(\beta, \xi) = \operatorname{LBS}(T_\text{cnl}(\beta, \xi), \mathcal{J}(\beta), \xi, \mathcal{W}_\text{lbs}),
\label{eq:smplx_lbs}
\end{equation}
where $\mathcal{J}(\beta) \in \mathbb{R}^{N_\text{j} \times 3}$ denotes the corresponding joint positions, and $\mathcal{W}_\text{lbs} \in \mathbb{R}^{N_\text{v} \times N_\text{j}}$ is a set of blend weights.

\subsection{SkelSD: Skeleton-Guided Score Distillation}\label{sec:method_skeleton_guided_sds}

Vanilla score distillation methods~\cite{dreamfusion, sjc} utilize view-dependent prompt augmentations such as ``front view of ...'' for diffusion model to provide crucial 3D view-consistent supervision. However, this prompting strategy cannot guarantee precise view consistency, leaving the disparity between the viewpoint of the diffusion model's supervision image and the 3D avatar's rendering image unresolved. Such inconsistency causes quality issues for 3D generation, such as blurriness and the Janus (multi-face) problem.

\begin{figure}[htbp]
\centering
\includegraphics[width=0.9\linewidth]{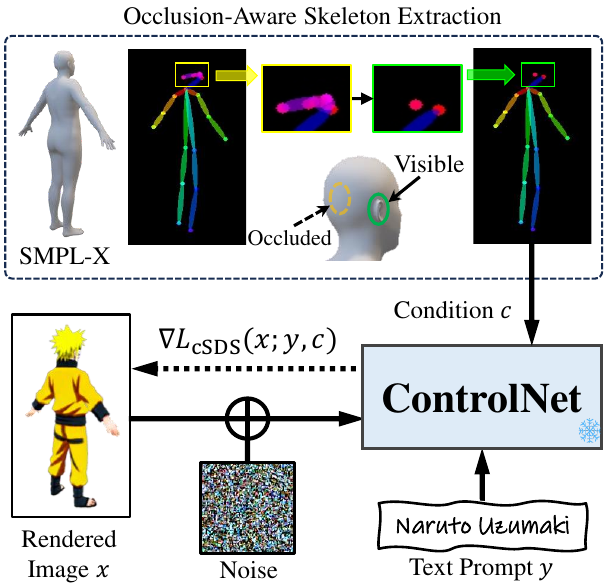}
\caption{The proposed \textbf{skeleton-guided score distillation} utilizes 2D skeleton images $c$ extracted from SMPL-X~\cite{smplx} to condition controllable 2D diffusion model (where we adopt ControlNet~\cite{controlnet}), which enhances the view and pose consistencies between the rendered image $x$ and the SDS supervision $\Delta L_\text{cSDS}$. In addition, we introduce occlusion culling to eliminate keypoints that are invisible from the current viewpoint, preventing ambiguity for the diffusion model.}
\label{fig:method:cSDS}
\end{figure}

\textbf{Skeleton-guided Score Distillation (SkelSD).} Inspired by recent works in controllable image generation~\cite{controlnet,humansd}, we propose \textbf{SkelSD}, which utilizes additional 3D-aware skeleton images from 3D human template~\cite{smplx} to condition SDS for view/pose-consistent score distillation, as shown in Figure~\ref{fig:method:cSDS}. Specifically, the skeleton conditioning image $c$ is injected to Equation~\ref{eq:sds} for SDS gradients, yielding:
\begin{equation*}
\quad\nabla_{\boldsymbol{\theta}}\mathcal{L}_{\text{cSDS}}(\phi,\mathbf{x})=\mathbb{E}_{t, \boldsymbol{\epsilon}}\bigg[w(t)
(\boldsymbol{\epsilon}_\phi(\mathbf{x}_t;y,t,{c})-\boldsymbol{\epsilon})\dfrac{\partial \mathbf{x}_t}{\partial\mathbf{x}}\dfrac{\partial \mathbf{x}}{\partial\boldsymbol{\theta}}\bigg],
\end{equation*}
where the conditioning image $c$ can be one or a combination of skeletons, depth maps, normal maps, etc. In practice, we opt for skeletons as the conditioning type because they offer minimal human shape priors, thereby facilitating the generation of complex geometries, as illustrated in Figure~\ref{fig:exp:grad_viz}. In order to acquire 3D-aware skeleton images, we use the parametric 3D human model SMPL-X~\cite{smplx} for skeleton rendering, where the skeleton image's viewpoint is strictly aligned with the avatar's rendering viewpoint.

\textbf{Occlusion Culling.} The introduction of 3D-aware conditioning images can enhance the 3D consistency in the SDS optimization process. However, the effectiveness is constrained by the adopted diffusion model~\cite{controlnet} on its interpretation of the conditioning images.
As shown in Fig.~\ref{fig:exp:occlusion_culling} (a), we provide a back-view skeleton map as the conditioning image to ControlNet~\cite{controlnet} and perform text-to-image generation. However, a frontal face still appears in the generated image. Such defects bring problems such as multiple faces (the Janus problem) and unclear facial features to 3D avatar generation. To this end, we propose to use occlusion culling algorithms~\cite{pantazopoulos2002occlusion} in computational graphics to detect whether facial keypoints are visible from the given viewpoint and subsequently remove them from the skeleton map if considered invisible. Body keypoints remain unaltered because they reside in the SMPL-X mesh, and it is difficult to determine whether they are occluded without introducing new priors.

\subsection{H3GA: Hybrid 3D Gaussian Avatars}\label{sec:method_hybrid_3dgs_avatar}

\begin{figure*}[htbp]
\centering
\includegraphics[width=\linewidth]{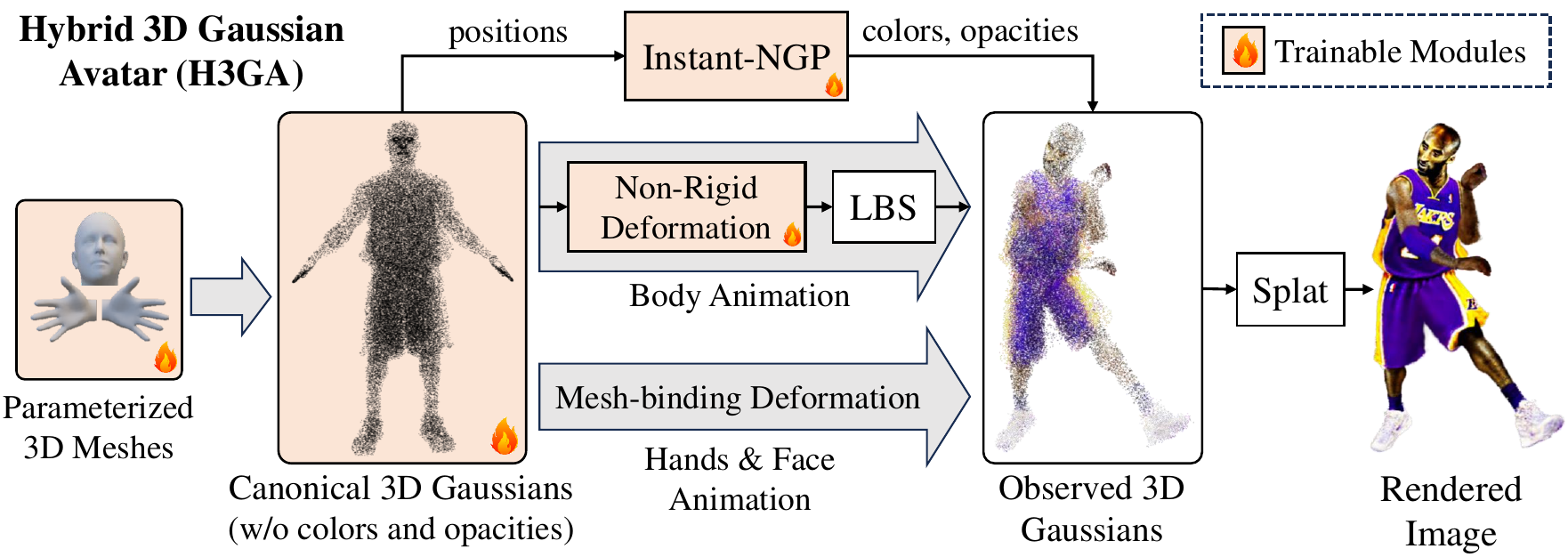}
\caption{The proposed \textbf{hybrid 3D Gaussian avatar representation} integrates efficient 3D Gaussian Splatting~\cite{3dgs} with neural implicit field (where we adopt Instant-NGP~\cite{instant-ngp}) and parameterized 3D meshes of SMPL-X~\cite{smplx} body parts (e.g., hands and face). Specifically, the canonical 3D Gaussian avatar is jointly represented by unconstrained 3D Gaussians $\Gu$ and mesh-binding 3D Gaussians $\Gm$ bound to parameterized 3D meshes. The colors and opacities of both $\Gu$ and $\Gm$ are predicted by the neural implicit field. For animation, $\Gu$ and $\Gm$ are deformed separately and merged to form observed 3D Gaussians, then splatted to obtain the rendered avatar image.}
\label{fig:method:H3GA}
\end{figure*}

The previous method DreamWaltz~\cite{huang2023dreamwaltz} utilizes NeRF~\cite{nerf} to represent 3D avatars, which is computationally expensive and results in extremely slow rendering and animation at high image resolutions (e.g., $1024\times1024$). To achieve higher training and inference efficiency, we adopt 3D Gaussian Splatting~\cite{3dgs} as the representation for 3D avatars.

Specifically for diffusion-guided 3D avatar creation, we review existing 3D Gaussian avatar representations~\cite{liu2024humangaussian,yuan2024gavatar} and propose several effective improvements for better generation and animation quality:
\begin{enumerate}
\item The high variance of score distillation gradients makes optimizing millions of 3D Gaussians challenging, as illustrated in Figure~\ref{fig:exp:ablation_H3GA}. Thus, we use pre-trained Instant-NGP~\cite{instant-ngp} to initialize the 3D Gaussians and to predict the 3D Gaussian properties for stable SDS optimization.
\item Considering that existing pre-trained 2D diffusion models struggle to generate intricate hands or control facial expressions, we embed the learnable 3D meshes of SMPL-X body parts (i.e., hands and face) into 3D Gaussians to ensure accurate geometry and animation for these body parts.
\item To articulate 3D Gaussians for animation, we bind each 3D Gaussian to the SMPL-X joints by assigning LBS weights and propose a geometry-aware smoothing algorithm based on K-Nearest Neighbors (KNN) for adaptive adjustments.
\item We introduce a deformation network conditioned on human pose to predict the pose-dependent variations of 3D Gaussian properties.
\end{enumerate}
These improvements constitute the proposed hybrid 3D Gaussian avatar representation, an overview of which is illustrated in Figure~\ref{fig:method:H3GA}.
 
\textbf{Formulation.} The proposed hybrid 3D Gaussian avatar representation consists of two types of 3D Gaussians: \( \mathcal{G}_\text{avatar} = \Gu \cup \Gm \), where \( \Gu \) denotes unconstrained 3D Gaussians, and \( \Gm \) denotes mesh-binding 3D Gaussians.

For unconstrained 3D Gaussians $\Gu$, the initial positions are extracted from a pre-trained NeRF. Specifically, we query NeRF to obtain the density distribution of a high-resolution 3D grid, and positions where the density exceeds a constant threshold are used as the initial positions $\mathbf{p}_u$ for $\Gu$. Then, the colors $\mathbf{c}_u$ and opacities $\alpha_u$ of $\Gu$ are predicted by:
\begin{equation}
    \mathbf{c}, \alpha = \text{NeRF} (\mathbf{p}).
\label{eq:gs_predict}
\end{equation}
The scales $\mathbf{s}_u$ and rotations $\mathbf{q}_u$ of $\Gu$ are explicitly initialized following 3DGS~\cite{3dgs} rather than being predicted by NeRF.

For mesh-binding 3D Gaussians $\Gm$, we utilize the pre-defined 3D meshes of the hands and face from SMPL-X and construct mesh-binding 3D Gaussians following SuGaR~\cite{guedon2024sugar} and GaMeS~\cite{waczynska2024games}. Exceptionally, the colors $\mathbf{c}_m$ and opacities $\alpha_m$ of $\Gm$ are predicted by NeRF following Equation~\ref{eq:gs_predict}. Besides, we parameterize the pre-defined 3D meshes using the shape parameters $\beta$ of SMPL-X, which are learnable.

\textbf{Articulation and Pose Transformation.} SMPL-X utilizes linear blend skinning (LBS)~\cite{mohr2003building} for the pose transformation of an articulated human body. This technique transforms the vertices of 3D meshes by blending multiple joint transformations based on LBS weights.
Therefore, for mesh-binding 3D Gaussians $\Gm$ bound to SMPL-X body parts, we can animate them by transforming the mesh vertices, following Equation~\ref{eq:smplx_lbs}.
For unconstrained 3D Gaussians $\Gu$, the pose transformation involves translating the position $\mathbf{p}$ and rotating the quaternion $\mathbf{q}$. We extend the LBS transformation of SMPL-X vertices to unconstrained 3D Gaussians as follows:
\begin{equation}
\Gu(\xi) = \operatorname{LBS}(\Gu^\text{cnl}, \mathcal{J}, \xi, \mathcal{W}_\text{lbs}),
\end{equation}
where $\Gu^\text{cnl}$ denotes unconstrained 3D Gaussians in the canonical pose, $\mathcal{J}$ represents SMPL-X joint positions, $\xi$ is the SMPL-X pose, and $\mathcal{W}_\text{lbs}$ is a set of LBS weights for $\Gu$. The acquisition of LBS weights $\mathcal{W}_\text{lbs}$ is given in Section~\ref{sec:method_lbs_init}.

\textbf{Non-rigid Deformation.} Pose-dependent deformations (i.e., $B_{P}(\xi)$ in Equation~\ref{eq:smplx_cnl}) allow the SMPL-X model to finely adjust and deform the body surface during pose changes. Still, it struggles to generalize to clothed avatars generated from texts. Thus we introduce a MLP-based deformation network~\cite{yang2024deformable} to model pose-dependent deformations for unconstrained 3D Gaussians $\Gu$:
\begin{equation}
    (\delta \mathbf{p}, \delta \mathbf{s}, \delta \mathbf{q}) = \operatorname{NRDeform} (\xi),
\end{equation}
where $(\delta \mathbf{p}, \delta \mathbf{s}, \delta \mathbf{q})$ represents the offsets of positions, scales, and quaternions of the unconstrained 3D Gaussians $\Gu^\text{cnl}$ in the canonical pose.
Note that the deformation network is subject-specific and trained from the diffusion guidance.

In addition, for mesh-binding 3D Gaussians $\Gm$, we model pose-dependent deformations following the mesh transformations of SMPL-X as described in Equation~\ref{eq:smplx_cnl}.

\subsection{DreamWaltz-G: Learning 3D Gaussian Avatars via Skeleton-guided Score Distillation}\label{sec:method_dreamwaltzG}

\begin{figure*}[htbp]
\centering
\includegraphics[width=\linewidth]{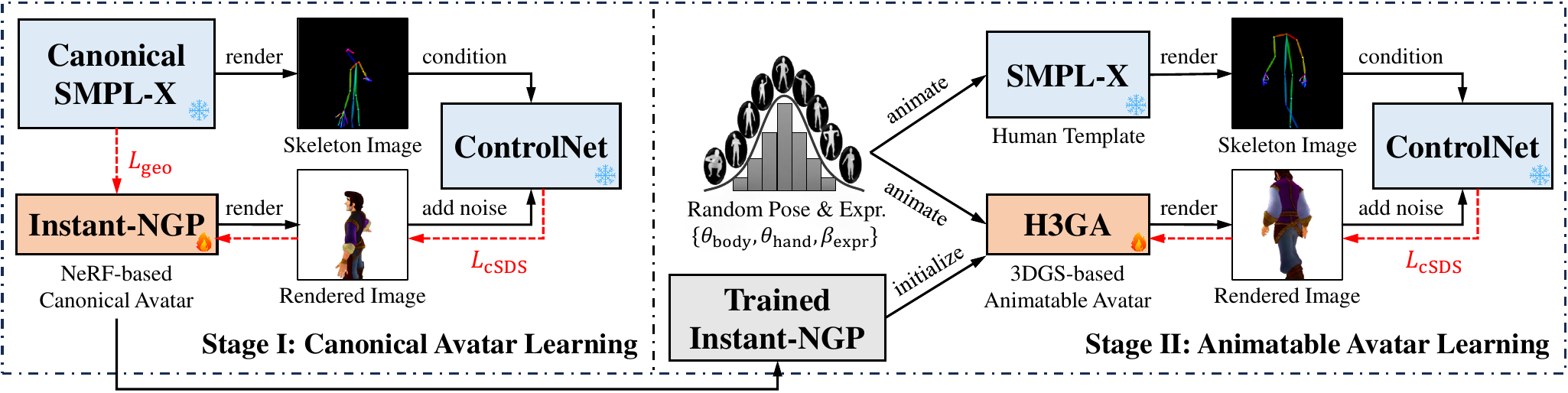}
\caption{The proposed animatable 3D avatar generation framework \textbf{DreamWaltz-G} consists of two training stages: \textbf{(I) Canonical Avatar Learning} and \textbf{(II) Animatable Avatar Learning}. In Stage I, We adopt the static Instant-NGP~\cite{instant-ngp} as canonical avatar representation. For each iteration, we extract a skeleton image from canonical SMPL-X~\cite{smplx} to condition ControlNet~\cite{controlnet}. Skeleton-conditioned score distillation loss $L_\text{cSDS}$ is used as a training objective to learn the canonical avatar. In Stage II, the proposed animatable avatar representation \textbf{H3GA} is first initialized with the trained Instant-NGP from Stage I and then optimized by $L_\text{cSDS}$. Unlike Stage I, which uses a fixed canonical pose, in Stage II, we randomly sample plausible human poses and expressions in each iteration to drive H3GA and SMPL-X, encouraging avatar learning across different motions.}
\label{fig:method:DreamWaltz-G}
\end{figure*}

Based on the proposed {Skeleton-guided Score Distillation} and {Hybrid 3D Gaussian Avatar Representation}, We further introduce a text-driven avatar generation framework: \textbf{DreamWaltz-G}. The framework comprises two training stages: (I) Static NeRF-based \textbf{{Canonical Avatar Learning}} (Sec.~\ref{sec:method_canonical_avatar}), (II) Deformable 3DGS-based \textbf{{Animatable Avatar Learning}} (Sec.~\ref{sec:method_animatable_avatar}), as illustrated in Figure~\ref{fig:method:DreamWaltz-G}.

\subsubsection{Canonical Avatar Learning}\label{sec:method_canonical_avatar}
In this stage, we employ a static NeRF (implemented with Instant-NGP~\cite{instant-ngp}) as the canonical avatar representation and train it using the skeleton-conditioned ControlNet~\cite{controlnet} and the canonical-posed SMPL-X model~\cite{smplx}. In particular, it leverages the SMPL-X model in three ways: (1) pre-training NeRF, (2) providing geometry constraints, and (3) rendering skeleton images to condition ControlNet for 3D-consistent and pose-aligned score distillation.

\textbf{Pre-training with SMPL-X.} To speed up the NeRF optimization and to provide reasonable initial renderings for the diffusion model, we pre-train NeRF based on an SMPL-X mesh template. Specifically, we render the silhouette and depth images of NeRF and SMPL-X given a randomly sampled viewpoint, and minimize the MSE loss between the NeRF renderings and the SMPL-X renderings.
The NeRF initialization from the human template significantly improves the geometry and the convergence efficiency for subsequent text-specific avatar generation.

\textbf{Score Distillation in Canonical Pose.} Given the target text prompt, we optimize the pre-trained NeRF through skeleton-guided score distillation loss $L^\text{cnl}_\text{cSDS}$ in the canonical pose space. We adopt the A-pose as the canonical pose because it best aligns with the diffusion prior and avoids leg overlap. Unlike  DreamWaltz~\cite{huang2023dreamwaltz} using SMPL~\cite{smpl} skeletons as condition images, we employ the more advanced SMPL-X~\cite{smplx} skeletons with hand joints and facial landmarks.

\textbf{Local Geometric Constraints of Body Parts.} During NeRF training, we introduce a local geometry loss based on pre-defined meshes of body parts, such as hands and faces. This ensures the trained NeRF is geometrically compatible with mesh-binding 3D Gaussians when serving as 3DGS initialization in subsequent stages.
Specifically, we align the NeRF densities $\tau$ of local regions with the pre-defined meshes using a margin ranking loss:
\begin{equation*}
L_\text{geo} = 
\begin{cases}
    (\max (0, \tau_\text{max} - \tau(\mathbf{p})) )^2 & \text{if}\ \mathbf{p}\ \text{on mesh} \\
    (\max (0, \tau(\mathbf{p}) - \tau_\text{min}) )^2 & \text{if}\ \mathbf{p}\ \text{not on mesh},
\end{cases}
\end{equation*}
where $\mathbf{p}$ represents 3D points sampled on and near the pre-defined meshes, $\tau(\mathbf{p})$ denotes the densities of 3D points $\mathbf{p}$ predicted by NeRF, $\tau_\text{min}$ and $\tau_\text{max}$ are constant hyperparameters.
Notably, Latent-NeRF~\cite{latentnerf} also introduces shape guidance to constrain NeRF geometry given a mesh sketch. Although both methods use pre-defined meshes as geometry guidance for NeRF optimization, the difference lies in their aim to provide a coarse geometry alignment, whereas we enforce strictly consistent geometries.

\textbf{Overall Objective.} To learn a canonical 3D avatar given text prompts, we optimize the NeRF-based static avatar representation using:
\begin{equation*}
L_\text{total}^\text{cnl} = L^\text{cnl}_\text{cSDS} + \lambda_\text{geo} L_\text{geo},
\end{equation*}
where $L^\text{cnl}_\text{cSDS}$ denotes the conditional SDS loss with canonical skeleton images as conditions, and $\lambda_\text{geo}=1.0$ is a balanced weight of the local geometry constraint.

\subsubsection{Animatable Avatar Learning}\label{sec:method_animatable_avatar}
In this stage, we initialize the proposed hybrid 3D Gaussians $\mathcal{G}_\text{avatar}$ as the animatable avatar representation and optimize it in random pose space using score distillation conditioned on SMPL-X skeletons.

\textbf{LBS Weight Initialization with SMPL-X.} \label{sec:method_lbs_init} Assigning LBS weights from SMPL-X vertices to each unconstrained 3D Gaussian $G \in \Gu$ is necessary for articulation and pose transformation. A naive implementation is mapping LBS weights based on nearest vertex criteria; however, this method cannot handle the geometric mismatches between SMPL-X and the generated avatars, leading to erroneous skeletal binding and distortions, as demonstrated in Figure~\ref{fig:exp:ablation_lbs}.
To address this, we propose using a geometry-aware KNN smoothing algorithm to adjust the assigned LBS weights of the 3D Gaussians adaptively. Specifically, for a 3D Gaussian $G \in \Gu$, its initial LBS weights $W^{(0)}_\text{lbs}$ can be derived from the nearest vertex in SMPL-X. Next, we update $W_\text{lbs}$ iteratively by weighted aggregation of the LBS weights $W_{\text{lbs},k}$ of the $K_\text{lbs}$ nearest 3D Gaussians:
\begin{equation}
    W^{(i+1)}_\text{lbs} = \sum_{k=1}^{K_\text{lbs}} \frac{Z_\text{lbs}}{ d_{\text{ng},k} \cdot d_{\text{nv},k} } W^{(i)}_{\text{lbs},k},
\end{equation}
where $i \in\{0, 1, \ldots, N_\text{lbs}\}$ denotes the current iteration step, $Z_\text{lbs}$ represents the normalization constant ensuring $Z_\text{lbs} \sum_{k=1}^{K_\text{lbs}} {(d_{\text{ng},k} \cdot d_{\text{nv},k})}^{-1} = 1$, $d_{\text{ng},k}$ is the squared distance from the $k$-th nearest 3D Gaussian $G_k$ to the current 3D Gaussian $G$, and $d_{\text{nv},k}$ is the squared distance from $G_k$ to its nearest vertex in SMPL-X. For clarity, $d_{\text{ng},k}^{-1}$ reflects the contribution of $G_k$ to $G$, while $d_{\text{nv},k}^{-1}$ indicates the confidence of the initial LBS weights of $G_k$.

\textbf{Score Distillation in Arbitrary Poses and Expressions.} Skeleton-guided score distillation $L^\text{arb}_\text{cSDS}$ in arbitrary poses helps to enhance visual quality and mitigate motion artifacts in novel poses. The previous work DreamWaltz~\cite{huang2023dreamwaltz} samples random poses using the off-the-shelf VPoser~\cite{smplx}, which is a variational autoencoder that learns a latent representation of human pose. However, optimizing directly in arbitrary pose spaces may be challenging to converge, leading to quality issues such as blurring. Therefore, we adopt a curriculum learning strategy from simple to difficult tasks, starting with sampling various canonical poses (such as A-pose, T-pose, and Y-pose), followed by sampling random poses from VPoser. Note that VPoser does not encompass hand poses and facial expressions. To obtain random hand poses and facial expressions, we randomly sample PCA coefficients from a Gaussian distribution and use the SMPL-X prior to compute corresponding pose and shape parameters.

\textbf{Overall Objective.} To learn an animatable 3D avatar given text prompts, we optimize the hybrid 3DGS-based dynamic avatar representation using $L^\text{arb}_\text{cSDS}$ only.
\section{Experiments}

\subsection{Implementation Details}\label{exp:implementation}
DreamWaltz-G is implemented in PyTorch and can be trained and evaluated on a single NVIDIA L40S GPU.

For the \textbf{Canonical Avatar Learning} stage, we employ Instant-NGP~\cite{instant-ngp} as the static 3D avatar representation. We optimize it for 15,000 iterations, which takes about one hour. We adopt a progressive resolution sampling strategy for efficient optimization, where the rendering resolution increases from 64$\times$64 to 512$\times$512 as iterations progress. More details on NeRF optimization, such as the optimizer and learning rate, are consistent with DreamWaltz~\cite{huang2023dreamwaltz}.

For the \textbf{Animatable Avatar Learning} stage, we use the proposed H3GA as the dynamic 3D avatar representation, which is trained for 15,000 iterations, and the rendering resolution is maintained at 512$\times$512. To optimize 3D Gaussian attributes, we adhere to the original implementation of 3DGS~\cite{3dgs}. However, we do not use the densification strategy for two reasons: (i) The high variance of SDS gradients makes gradient-based densification unstable; (ii) The initialization based on a trained NeRF can provide accurate and quantitative 3D Gaussians.

\textbf{Diffusion Guidance.} We use \textit{Stable-Diffusion-v1.5}~\cite{latentdiffusion} and \textit{ControlNet-v1.1-openpose}~\cite{controlnet} to provide SDS guidance for both training stages. We randomly sample the timestep from a uniform distribution of $[0.02, 0.98]$, and the classifier-free guidance scale is set to $50.0$. The weight term $w(t)$ for SDS loss is set to $1.0$. The conditioning scale for ControlNet is set to $1.0$ by default.
To further improve 3D consistency and visual quality, both view-dependent text augmentation~\cite{dreamfusion} and negative prompts are used.

\textbf{Camera Sampling.} For each iteration, the camera view is randomly sampled in spherical coordinates, where the radius, azimuth, elevation, and FoV are uniformly sampled from $[1.0,2.0]$, $[0,360]$, $[60,120]$, and $[40,70]$, respectively. The camera focus strategy is also employed, with a 0.2 probability of focusing on the face of the 3D avatar to enhance facial details. Additionally, we empirically find that horizontal camera jitter during training helps improve the visual quality of the foot region.

\textbf{Motion Sequences.} To create animation demonstrations, we utilize SMPL-X motion sequences from 3DPW~\cite{3dpw}, AIST++~\cite{aist}, Motion-X~\cite{motionx}, and TalkSHOW~\cite{talkshow} datasets to animate avatars. SMPL-X motion sequences extracted from in-the-wild videos are also used.

\begin{figure*}[htbp]
\centering
\includegraphics[width=\linewidth]{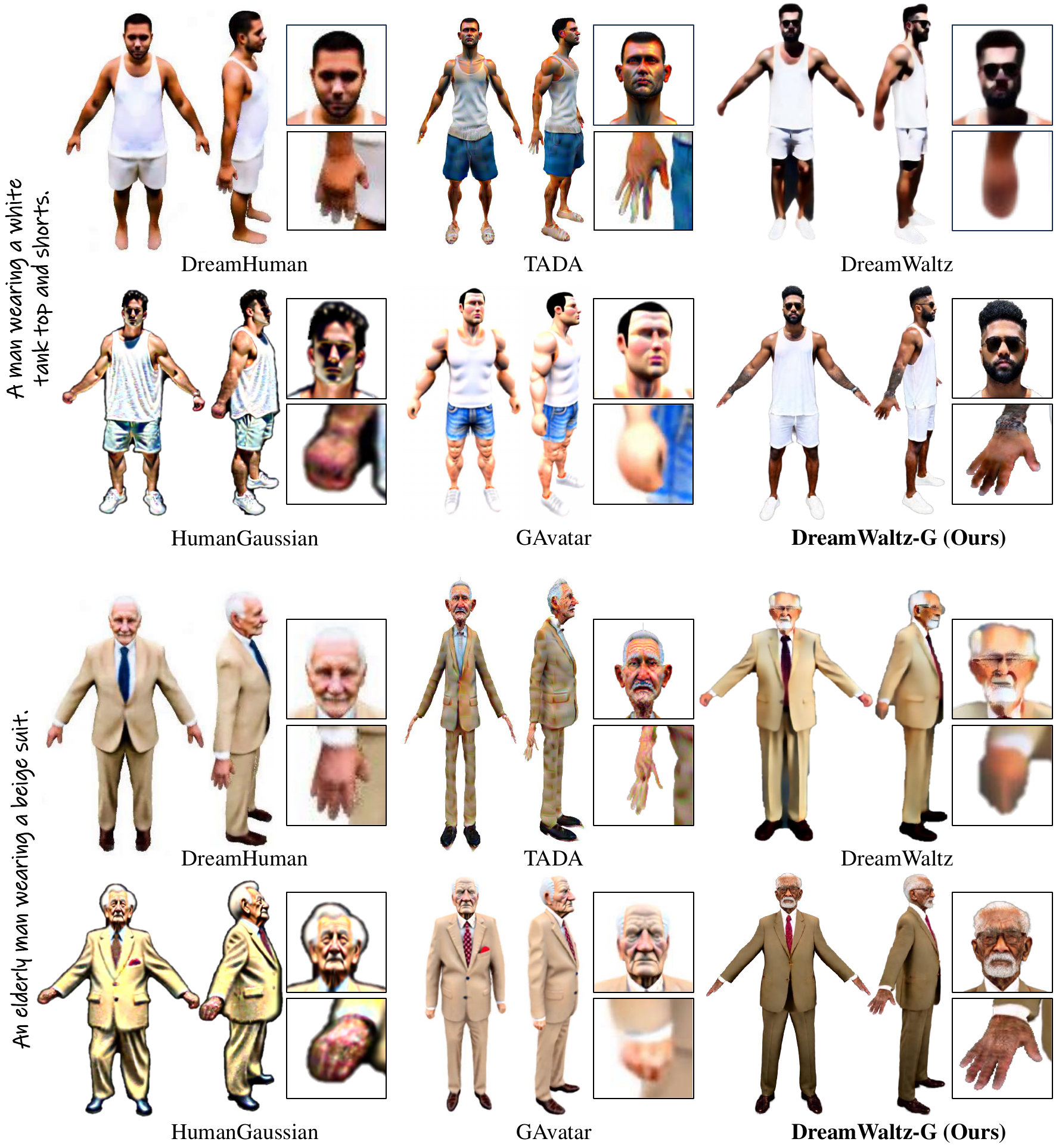}
\caption{\textbf{Qualitative results of canonical avatars} compared to existing text-driven 3D avatar generation methods: DreamWaltz~\cite{huang2023dreamwaltz}, DreamHuman~\cite{kolotouros2024dreamhuman}, TADA~\cite{liao2024tada}, GAvatar~\cite{yuan2024gavatar}, HumanGaussian~\cite{liu2024humangaussian}. The text prompts used are listed on the left.}
\label{fig:exp:comp_cnl}
\end{figure*}

\subsection{Comparisons}\label{exp:avatar_animation}
We provide both qualitative and quantitative results of our DreamWaltz-G compared to existing text-driven 3D avatar generation methods, including DreamWaltz~\cite{huang2023dreamwaltz}, DreamHuman~\cite{kolotouros2024dreamhuman}, TADA~\cite{liao2024tada}, HumanGaussian~\cite{liu2024humangaussian}, and GAvatar~\cite{yuan2024gavatar}.

\begin{figure*}[htbp]
\centering
\includegraphics[width=\linewidth]{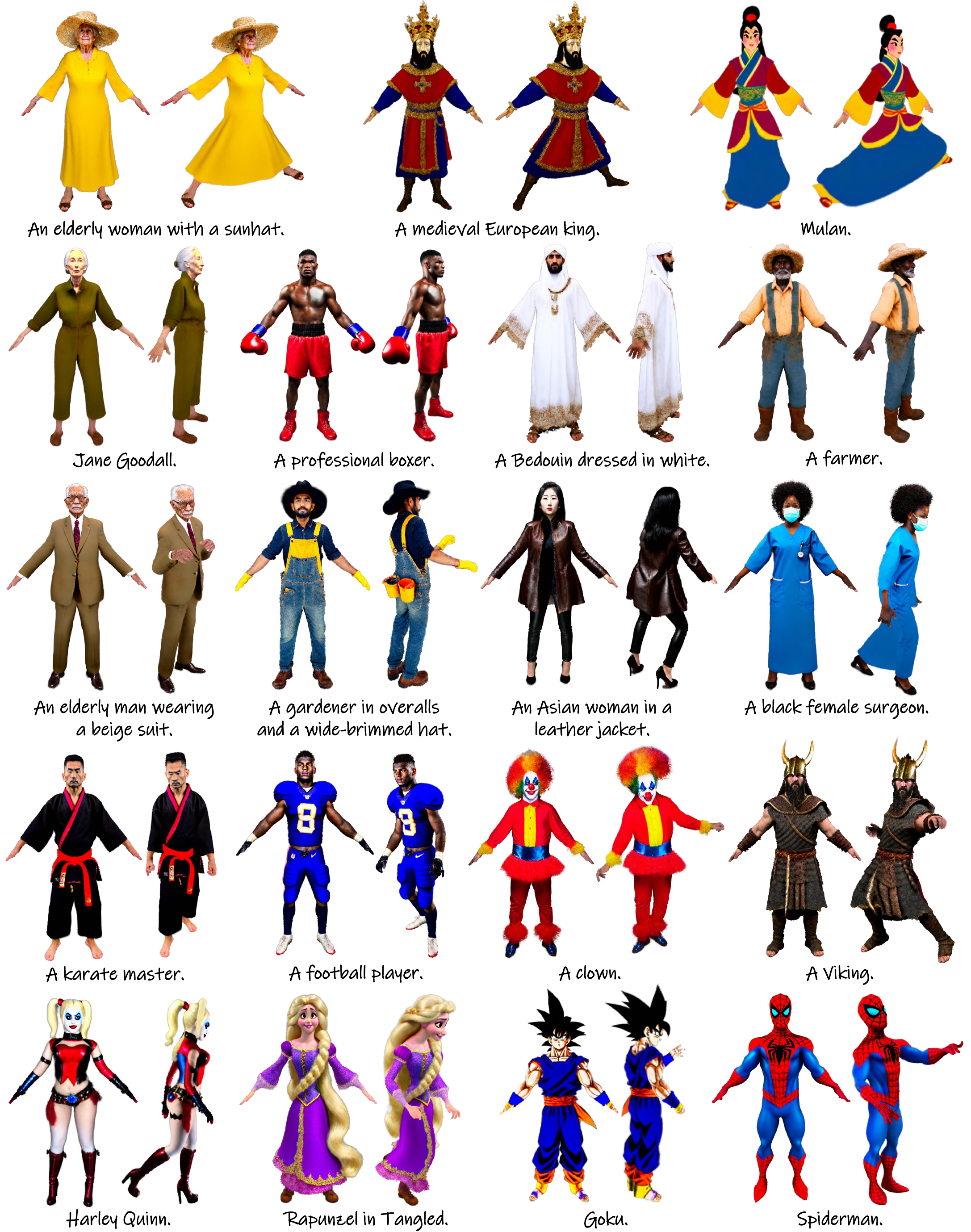}
\caption{\textbf{More examples of 3D avatars and their animations} produced by our approach. The text prompts used are listed below.}
\label{fig:exp:gallery}
\end{figure*}

\begin{figure*}[htbp]
\centering
\includegraphics[width=0.94\linewidth]{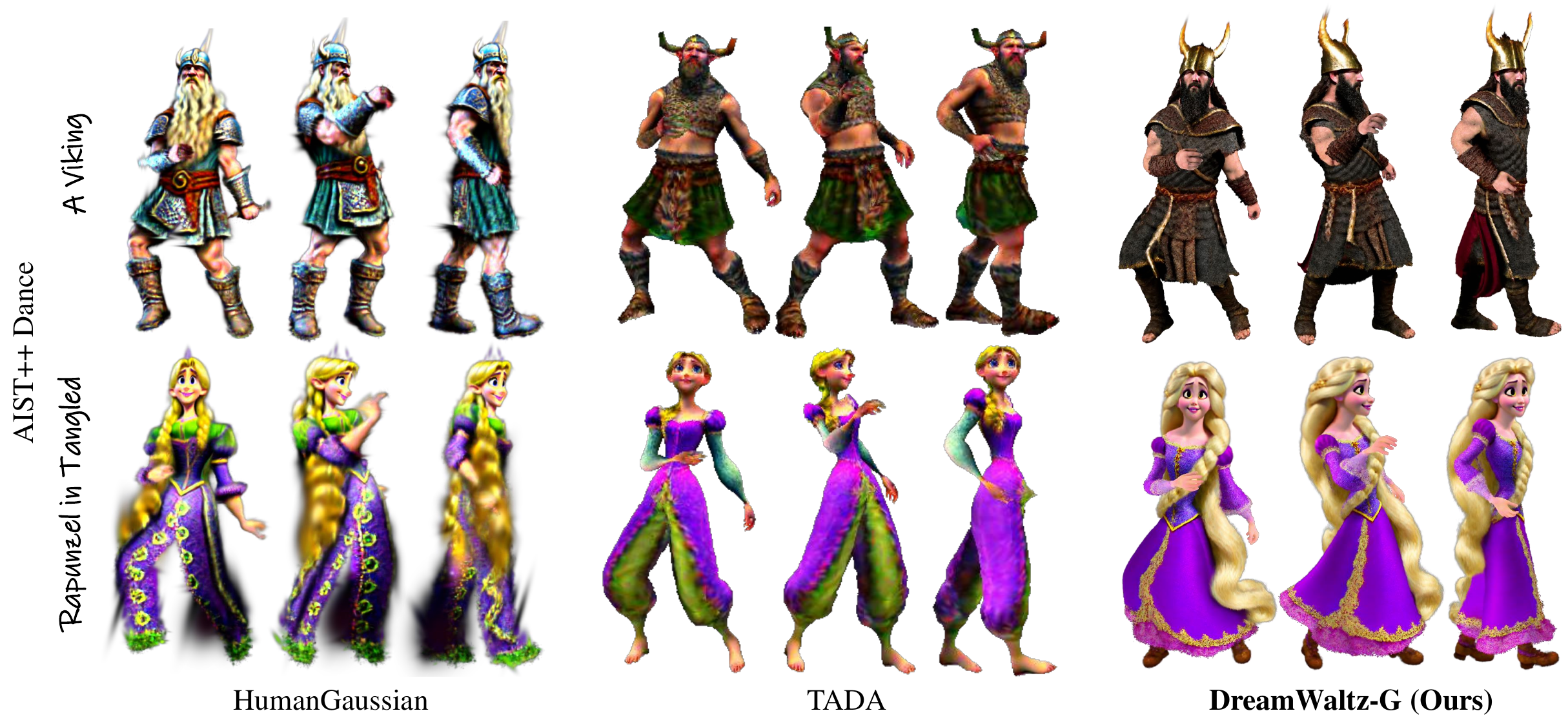}
\caption{\textbf{Qualitative results of animatable avatars} compared to existing 3d avatar generation and animation methods: HumanGaussian~\cite{liu2024humangaussian} and TADA~\cite{liao2024tada}. Compared to competing methods, our approach achieves clearer hand motions and higher-fidelity animation quality. In comparison to HumanGaussian, which is also based on 3DGS~\cite{3dgs}, we effectively avoid sharp artifacts caused by the incorrect driving of 3D Gaussians.}
\label{fig:exp:comp_ani}
\end{figure*}

\begin{table*}[htbp]
    \centering
    \caption{\textbf{User preference studies.} We report the preference percentages (\%) of our method over existing state-of-the-art methods in terms of geometric quality, appearance quality, and consistency with the text prompts.}
    \renewcommand\arraystretch{1.2}
    \begin{tabular}{l|c|c|c}
        \hline
        \textbf{Methods} & \textbf{Geometry Quality} & \textbf{Appearance Quality} & \textbf{Text Consistency} \\
        \hline
        Ours vs. DreamWaltz~\cite{huang2023dreamwaltz} & 84.93 & 86.30 & 78.08 \\
        Ours vs. DreamHuman~\cite{kolotouros2024dreamhuman} & 82.61 & 86.96 & 84.78 \\
        Ours vs. TADA~\cite{liao2024tada} & 70.27 & 77.03 & 66.22 \\
        Ours vs. GAvatar~\cite{yuan2024gavatar} & 82.05 & 76.92 & 79.49 \\
        Ours vs. HumanGaussian~\cite{liu2024humangaussian} & 70.31 & 75.00 & 76.56 \\
        \hline
    \end{tabular}
    \label{tab:user_study}
\end{table*}

\textbf{Qualitative Results of Canonical Avatars.} We present the results of canonical avatars, as shown in Figure~\ref{fig:exp:comp_cnl}. Compared to existing methods, our approach achieves high-definition and realistic appearances, alleviating blurriness and over-saturation issues. Additionally, our approach can generate accurate hand and facial shapes by leveraging the geometric priors of predefined meshes, addressing the diffusion model's difficulty in generating detailed human body parts. We provide more examples of canonical 3D avatars generated by our method in Figure~\ref{fig:exp:gallery}.

\begin{figure*}[htbp]
\centering
\includegraphics[width=1.\linewidth]{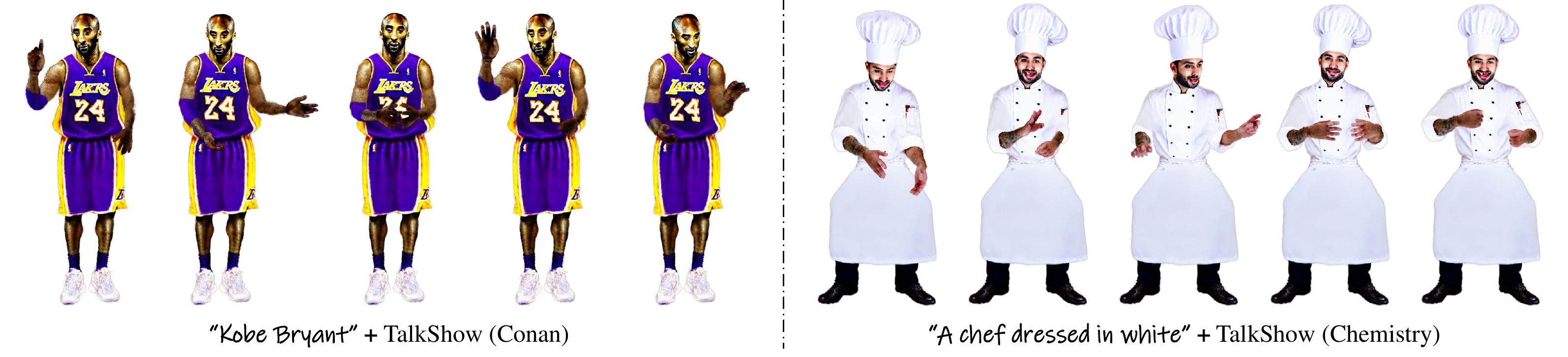}
\caption{\textbf{Application: Talking 3D Avatars.} Benefiting from the proposed expressive H3GA representation, our method can learn animatable 3D avatars from 2D diffusion priors while preserving the fine details of hands and faces. This allows us to create more expressive 3D avatar animations like talking 3D avatars.}
\label{fig:exp:app_talkshow}
\end{figure*}

\textbf{Qualitative Results of Animatable Avatars.} We demonstrate the animation results of our method compared to HumanGaussian~\cite{liu2024humangaussian} and TADA~\cite{liao2024tada}, as shown in Figure~\ref{fig:exp:comp_ani}. The SMPL-X motion sequences from the AIST++ dance dataset~\cite{aist} are used to animate the generated avatars. Compared to existing competing methods, our approach achieves clearer hand motions and higher-fidelity animation quality. In comparison to HumanGaussian, which is also based on 3DGS~\cite{3dgs}, we effectively avoid sharp artifacts caused by the incorrect driving of 3D Gaussians. More examples of avatar animations can be seen in Figure~\ref{fig:exp:gallery} and Figure~\ref{fig:exp:app_talkshow}.

\textbf{User Studies.} To quantitatively evaluate the quality of the generated 3D avatars compared to existing methods, we conducted a A/B user preference study based on 24 text prompts released by GAvatar~\cite{yuan2024gavatar}. 
Twenty participants are asked to view 3D avatars generated by our method and one of the competing methods and then choose the better method based on (1) geometric quality, (2) appearance quality, and (3) consistency with the text prompts. As reported in Table~\ref{tab:user_study}, the participants favor 3D avatars generated by our method across all evaluation criteria.

\subsection{Ablation and Analysis}\label{exp:ablations}
We perform a comprehensive ablation analysis to demonstrate the effectiveness of the proposed improvements.

\begin{figure*}[htbp]
\centering
\includegraphics[width=1.\linewidth]{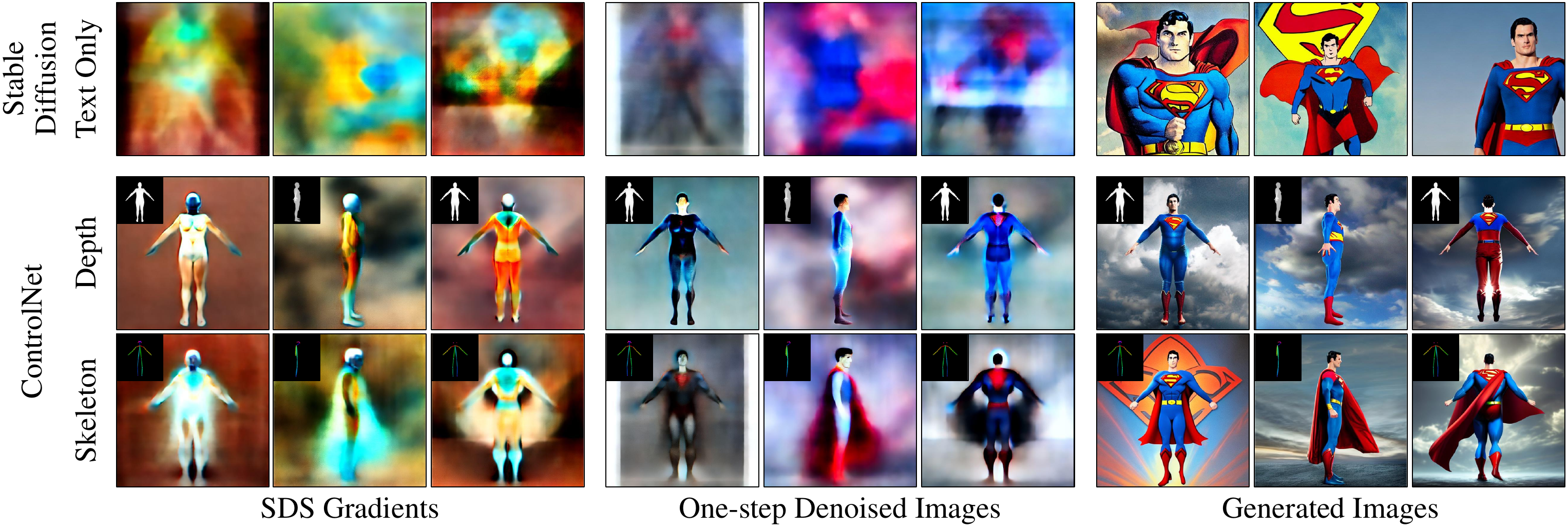}
\caption{\textbf{Visualization of SDS gradients and generated images under different guidance conditions.} The results in the first row are conditioned only on text. In contrast, the second and third rows are conditioned on additional depth and skeleton images, respectively, as indicated in the upper left corner of each visualization. These results are based on the text prompt ``superman''. It is evident that skeleton conditions, as adopted by our DreamWaltz-G, provide more informative supervision than text-only conditions. Skeleton conditions are also less restrictive than depth conditions, successfully avoiding the loss of complex appearances, such as the disappearance of Superman's cape.
}
\label{fig:exp:grad_viz}
\end{figure*}

\begin{figure*}[htbp]
\centering
\includegraphics[width=\linewidth]{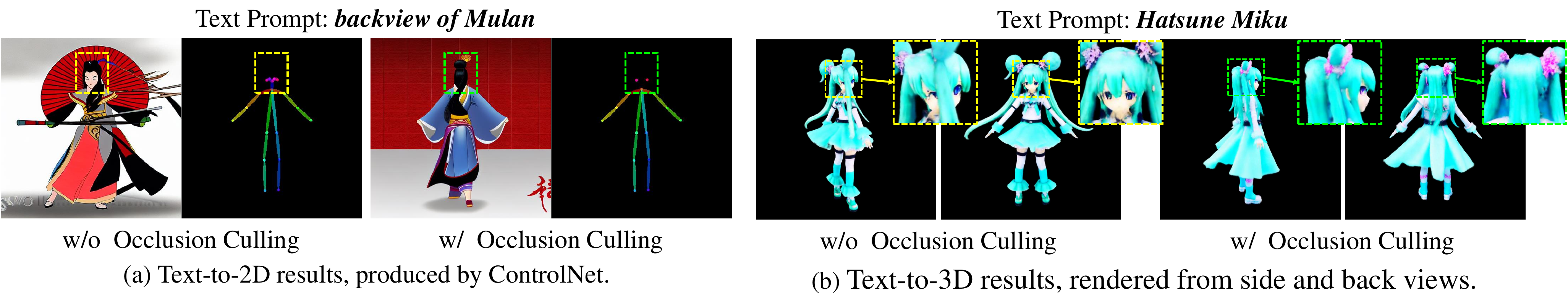}
\caption{\textbf{Ablation studies on occlusion culling.} We employ occlusion culling to refine skeleton condition images by removing invisible human keypoints, such as the eyes and nose in the back view. It helps (a) ControlNet~\cite{controlnet} to generate the character's back view correctly, and (b) text-to-3D avatar generation to resolve the multi-face problem, as highlighted by the bounding boxes.}
\label{fig:exp:occlusion_culling}
\end{figure*}

\textbf{Effectiveness of Skeleton Guidance.}
We visualize the SDS gradients and generated images in Figure~\ref{fig:exp:grad_viz} to illustrate the advantages of skeleton guidance compared to text-only guidance and depth guidance. It is evident that depth and skeleton images from human templates offer more informative guidance than text alone. However, the strong contour priors in depth images cause the SDS gradients to conform tightly to the avatar's skin, leading to a lack of complex appearances (e.g., the disappearance of Superman's cape in the second row of Figure~\ref{fig:exp:grad_viz}). On the other hand, skeleton images, as adopted by DreamWaltz-G, provide both informative and flexible supervision, accurately capturing the avatars' poses and intricate shapes.

\textbf{Ablation Studies on Occlusion Culling.}
Occlusion culling is crucial for resolving view ambiguity both for skeleton-conditioned 2D and 3D generation, as shown in Figure~\ref{fig:exp:occlusion_culling}. Limited by the view-aware capability, ControlNet~\cite{controlnet} fails to generate the back-view image of a character even with view-dependent text and skeleton prompts, as shown in Figure~\ref{fig:exp:occlusion_culling}(a). The introduction of occlusion culling eliminates the ambiguity of skeleton conditions and helps ControlNet to generate correct views. Similar effects can be observed in text-to-3D avatar generation. As shown in Figure~\ref{fig:exp:occlusion_culling}(b), The Janus (multi-face) problem is solved by introducing occlusion culling to the rendering process from 3D SMPL-X to the 2D skeleton images.

\begin{figure*}[htbp]
\centering
\includegraphics[width=1.\linewidth]{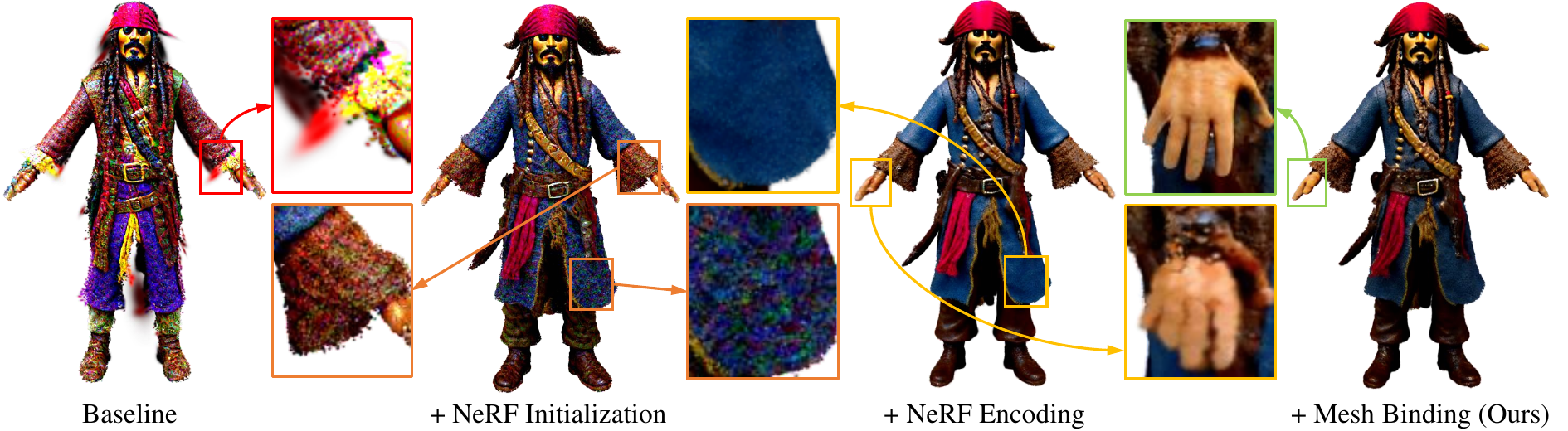}
\caption{\textbf{Ablation studies on the proposed {H}ybrid {3}D {G}aussian {A}vatar representation}, which incorporates several improvements to accommodate SDS optimization and enable expressive avatar animation. Specifically, ``NeRF Initialization'' provides a well-structured point cloud to initialize the 3D Gaussians, facilitating the capture of complex geometries. ``NeRF Encoding'' utilizes Instant-NGP~\cite{instant-ngp} to predict 3D Gaussian attributes, resulting in more stable SDS optimization and avoiding high-frequency noise in textures. For intricate body parts like hands, we adopt a “Mesh Binding” strategy, which binds the corresponding 3D Gaussians to the SMPL-X body parts, achieving sharp and joint-aligned geometries.}
\label{fig:exp:ablation_H3GA}
\end{figure*}

\begin{figure}[htbp]
\centering
\includegraphics[width=1.\linewidth]{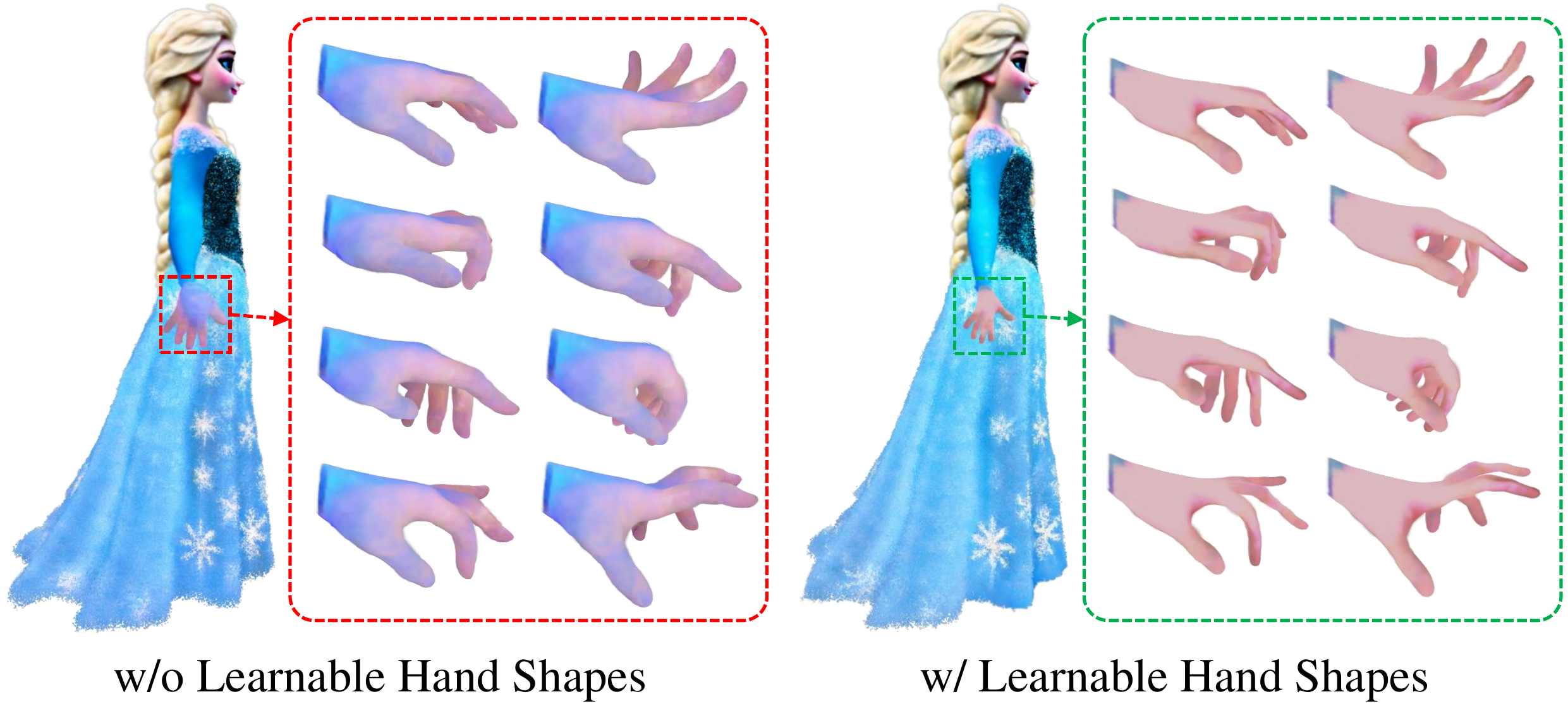}
\caption{\textbf{Ablation studies on learnable shape parameters} (e.g., $\beta_\text{hand}$ of SMPL-X~\cite{smplx}) for mesh-binding 3D Gaussian body parts. We use the hands of ``Princess Elsa in Frozen'' as an example to demonstrate. By optimizing the hand shape parameters of mesh-binding 3D Gaussians, slimmer hands that match Elsa's characteristics can be generated.}
\label{fig:exp:ablation_betas}
\end{figure}

\begin{figure}[htbp]
\centering
\includegraphics[width=0.95\linewidth]{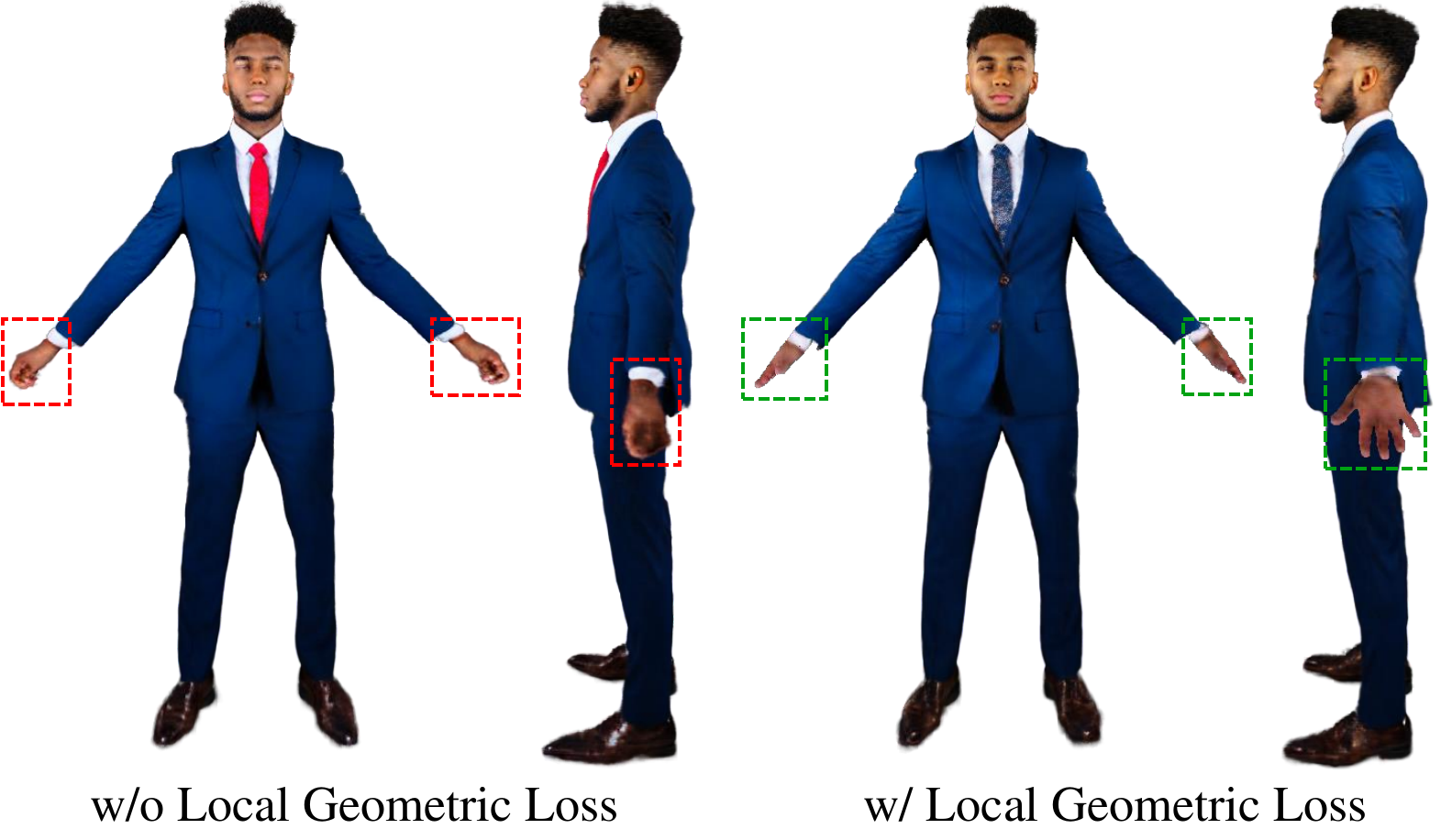}
\caption{\textbf{Ablation studies on local geometric constraints.} Without the local geometric loss $L_\text{geo}$, the generated avatar's hands appear in a clenched fist state (highlighted by {{dashed boxes}}), exhibiting unclear geometric structures. The introduction of $L_\text{geo}$ ensures that the hand structure is accurately aligned with canonical SMPL-X (highlighted by {{dashed boxes}}), avoiding erroneous geometries and facilitating subsequent rigging and hand animation.}
\label{fig:exp:ablation_lgc}
\end{figure}

\begin{figure}[htbp]
\centering
\includegraphics[width=.9\linewidth]{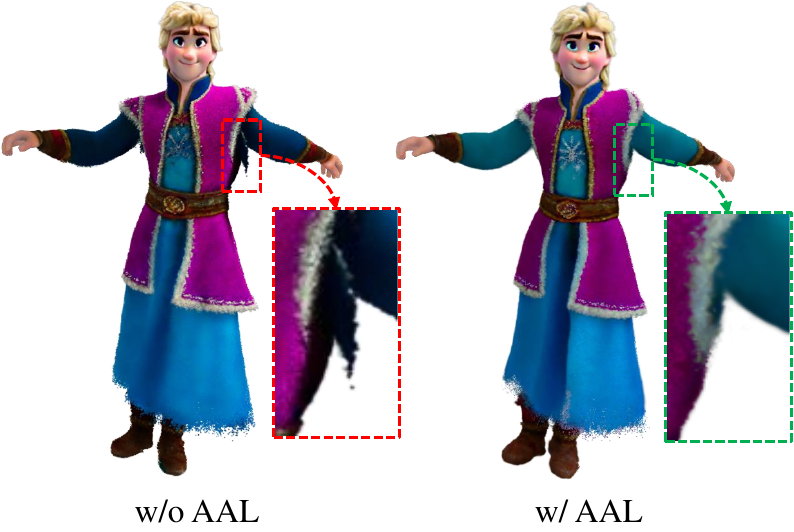}
\caption{\textbf{Ablation studies on Animatable Avatar Learning (AAL)}, which is the Stage II of DreamWaltz-G. For ``w/o AAL'', we train for the same iterations as ``w/ AAL'' but use a fixed canonical pose to ensure a fair comparison. It can be observed that the introduction of AAL fixes texture information for areas not visible in the canonical pose. Besides, it reduces animation artifacts caused by incorrect skeleton binding.}
\label{fig:exp:ablation_AAL}
\end{figure}

\begin{figure*}[htbp]
\centering
\includegraphics[width=1.\linewidth]{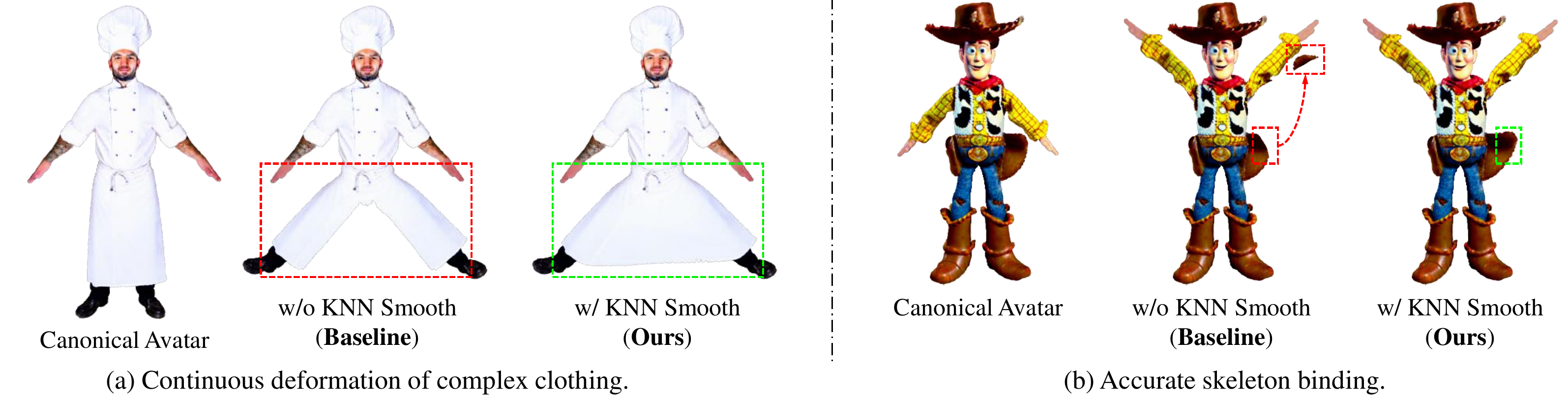}
\caption{\textbf{Ablation studies on KNN smoothing for LBS weight initialization.} The proposed geometry-aware KNN Smoothing algorithm refines the 3D Gaussians' initial LBS weights (representing the association of each 3D Gaussian to body joints). Compared to the baseline that assigns LBS weights based solely on the nearest neighbor criterion, the proposed algorithm enables (a) continuous deformation of complex clothing, e.g., the stretching of the chef's apron; (b) accurate skeleton binding, for example, the hat hanging from Woody’s waist is not affected by arm movements.}
\label{fig:exp:ablation_lbs}
\end{figure*}

\begin{figure*}[htbp]
\centering
\includegraphics[width=1.\linewidth]{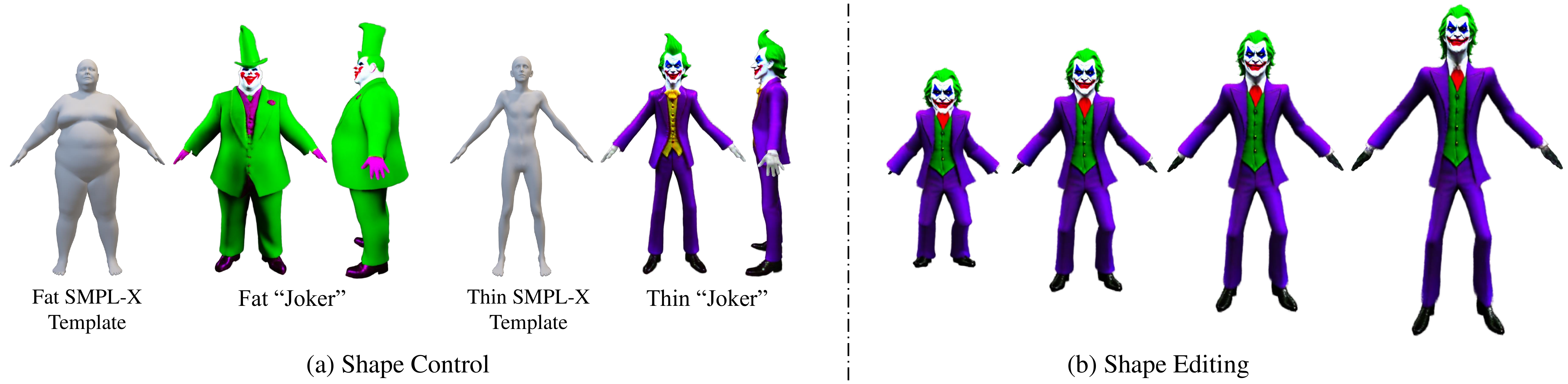}
\caption{\textbf{Application: Shape Control and Editing.} Our method enables (a) training-time shape control by modifying the SMPL-X template and (b) inference-time shape editing during inference by explicitly adjusting the 3D Gaussians. Both shape control and editing are compatible with the SMPL-X shape parameters $\beta$, allowing users to simply adjust $\beta$ to achieve the desired 3D shape.}
\label{fig:exp:app_shape}
\end{figure*}

\textbf{Ablation Studies on Hybrid 3D Gaussian Avatars.} The proposed 3D avatar representation, H3GA, incorporates several improvements to accommodate SDS optimization and enable expressive avatar animation. We analyze the effects of these improvements individually, as shown in Figure~\ref{fig:exp:ablation_H3GA}. Specifically, ``NeRF Initialization'' provides a well-structured point cloud to initialize the 3D Gaussians, facilitating the capture of complex geometries that differ from SMPL-X templates. ``NeRF Encoding'' utilizes multi-resolution hash grids~\cite{instant-ngp} and MLPs to predict 3D Gaussian attributes, resulting in more stable SDS optimization and avoiding high-frequency noise in textures.

For body parts that are challenging to generate and animate (e.g., hands and face), we adopt a “Mesh Binding” strategy. This strategy binds the corresponding 3D Gaussians to the meshes of SMPL-X body parts, achieving sharp and joint-aligned geometries. Note that these mesh-binding body parts are parameterized by SMPL-X shape parameters and are trainable. As shown in Figure~\ref{fig:exp:ablation_betas}, hands that conform to the character's features can be obtained by optimizing the SMPL-X hand shape parameters.

\textbf{Ablation Studies on Local Geometric Constraints.} The local geometric constraints $L_\text{geo}$ are introduced during canonical NeRF training to maintain the geometric structures of intricate body parts, such as hands and faces. As shown in Figure~\ref{fig:exp:ablation_lgc}, without the local geometric loss, the generated avatar's hands appear in a clenched fist state, exhibiting unclear geometric structures and difficulties with rigging and animation. Introducing the local geometric loss ensures that the hand structure is accurately aligned with canonical SMPL-X, avoiding erroneous geometries and facilitating subsequent hand animation.

\textbf{Ablation Studies on DreamWaltz-G.} The proposed avatar generation framework, DreamWaltz-G, consists of two training stages: Canonical Avatar Learning (CAL), and Animatable Avatar Learning (AAL). The CAL stage aims to provide a good NeRF initialization for H3GA, the effectiveness of which is validated as shown in Figure~\ref{fig:exp:ablation_H3GA}. The AAL stage aims to learn the appearance and geometry of the 3D avatar in a random pose space. As shown in Figure~\ref{fig:exp:ablation_AAL}, the introduction of AAL fixes texture information for areas not visible in the canonical pose and reduces animation artifacts caused by incorrect skeleton binding.

\textbf{Ablation Studies on KNN Smoothing for LBS Weight Initialization.} We propose a geometry-aware KNN Smoothing algorithm to refine the initial LBS weights (representing the association of each 3D Gaussian to body joints), bringing various improvements in avatar rigging and animation. As shown in Figure~\ref{fig:exp:ablation_lbs}, the proposed KNN smoothing algorithm enables: (a) continuous deformation of complex clothing, e.g., the stretching of a dress; (b) accurate skeleton binding, which should be geometry-aware rather than based solely on the nearest neighbor criterion.

\subsection{Applications}\label{exp:applications}

\begin{figure*}[htbp]
\centering
\includegraphics[width=1.\linewidth]{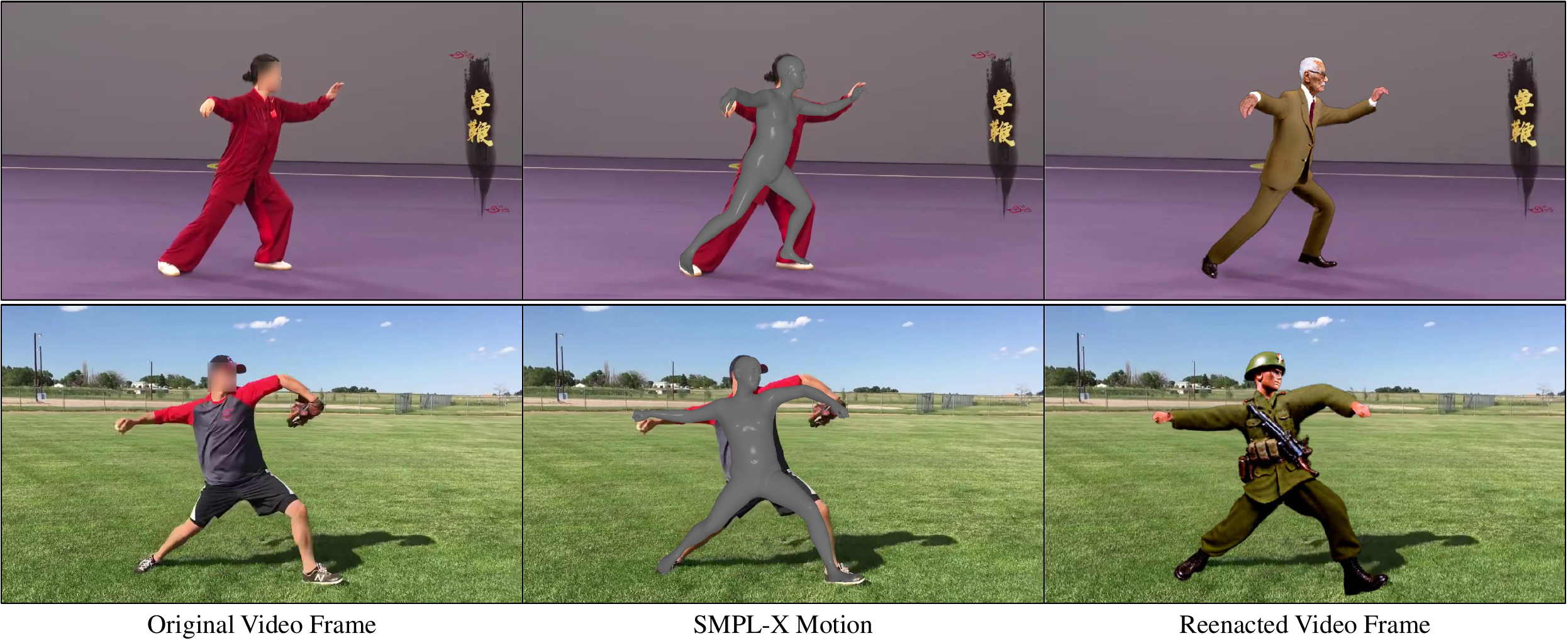}
\caption{\textbf{Application: Human Video Reenactment.} Combined with 3D human pose estimation and video inpainting techniques, the 3D avatars generated by our method can be projected onto 2D human videos. This integration allows for seamless blending of animated 3D avatars with real-world footage, enhancing the realism and interactivity of the reenacted scenes.}
\label{fig:exp:app_reenact}
\end{figure*}

\begin{figure*}[htbp]
\centering
\includegraphics[width=1.\linewidth]{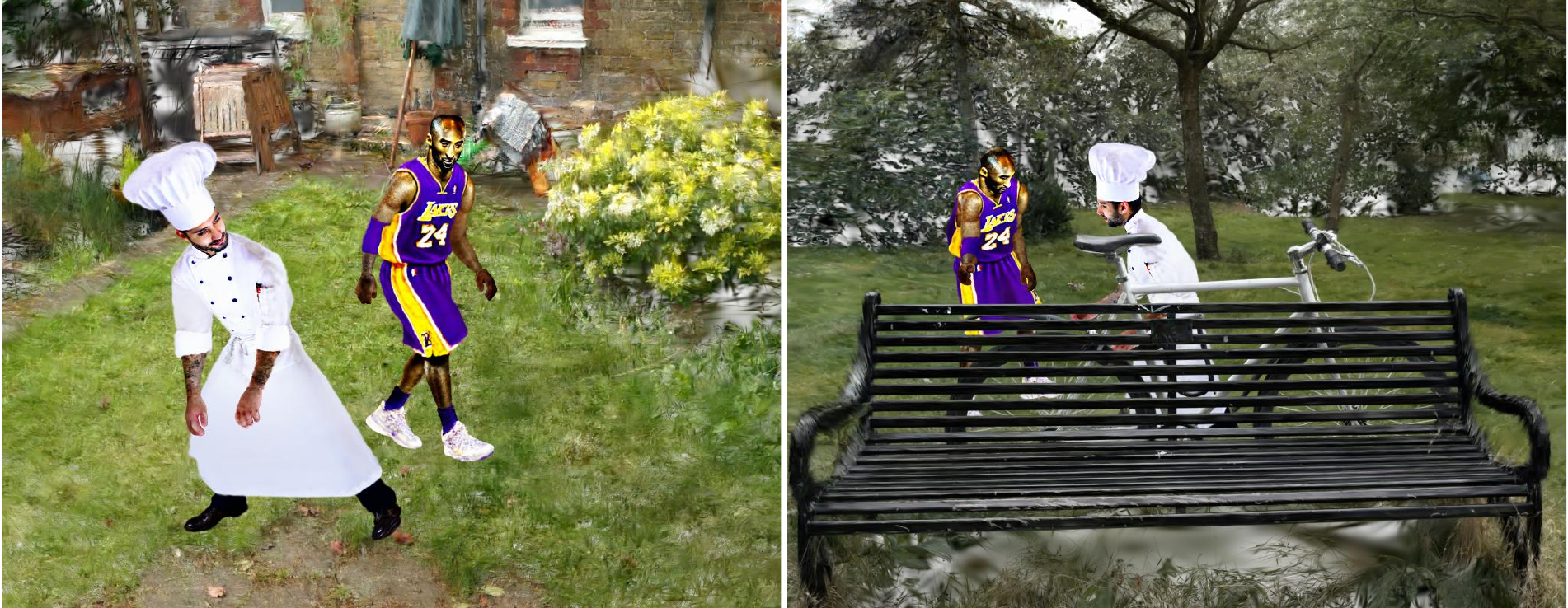}
\caption{\textbf{Application: Multi-subject Scene Composition.} The generated 3D avatars can be seamlessly integrated with existing 3D assets. The presented 3D environments are from the Mip-NeRF 360 dataset~\cite{mip-nerf360} and reconstructed by vanilla 3D Gaussian Splatting~\cite{3dgs}.}
\label{fig:exp:app_scene}
\end{figure*}

We explore practical applications of our method, including: shape control and editing, talking 3D avatars, human video reenactment, and multi-subject 3D scene composition.

\textbf{Shape Control and Editing.} Our method utilizes the SMPL-X template to provide skeleton guidance for 3D avatar creation. By adjusting the shape parameters of the SMPL-X template, the shape of the generated 3D avatar can be controlled, as shown in Figure~\ref{fig:exp:app_shape}(a). However, this shape control requires re-training, which leads to inefficiency and appearance randomness. Thanks to the explicit 3D avatar representation, our method can also achieve shape editing by adjusting the 3D Gaussians. Compared to shape control, shape editing is real-time, interactive, and able to maintain a consistent appearance, as shown in Figure~\ref{fig:exp:app_shape}(b).

\textbf{Talking 3D Avatars.} The proposed H3GA representation enables the modeling of animatable 3D avatars from 2D diffusion priors while preserving the fine details of hands and faces. This allows us to create more expressive 3D avatar animations, for example, talking 3D avatars. As shown in Figure~\ref{fig:exp:app_talkshow}, the results exhibit realistic appearances, intricate geometries, and accurate hand and face animations.

\textbf{Human Video Reenactment.} Combined with 3D human pose estimation~\cite{motionx} and video inpainting techniques, the 3D avatars generated by our method can be projected onto 2D human videos, as shown in Figure~\ref{fig:exp:app_reenact}. This integration allows for seamless blending of animated 3D avatars with real-world footage, enhancing the realism and interactivity of the reenacted scenes.

\textbf{Multi-subject Scene Composition.} The generated 3D avatars can be integrated with existing 3D assets into the same scene. As shown in Figure~\ref{fig:exp:app_scene}, we place the animated 3D avatars ``Kobe Bryant'' and ``a chef dressed in white'' into 3D scenes, seamlessly integrating the avatars into the environment.

\section{Conclusions}
We introduce DreamWaltz-G, a novel learning framework for animatable 3D avatar generation from texts. At the core of this framework are skeleton-guided score distillation and hybrid 3D Gaussian avatar representation. Specifically, we leverage the skeleton priors from the human parametric model~\cite{smplx} to guide the score distillation process, providing 3D-consistent and pose-aligned supervision for high-quality avatar generation. The hybrid 3D Gaussian representation builds on the efficiency of 3D Gaussian splatting~\cite{3dgs}, combining NeRF~\cite{nerf} and 3D meshes~\cite{guedon2024sugar} to accommodate SDS optimization and enable expressive animations. Extensive experiments demonstrate that DreamWaltz-G is effective and outperforms existing text-to-3D avatar generation methods in both visual quality and animation. Benefiting from DreamWaltz-G, we could unleash our imagination and enable a wide range of avatar applications.

Similar to previous 3D generation methods~\cite{dreamfusion,sjc,huang2023dreamwaltz}, DreamWaltz-G generates 3D avatars through score distillation~\cite{dreamfusion}. Leveraging more powerful foundational models~\cite{sdxl,sd3} and advanced score distillation techniques~\cite{dreamtime,nfsd} can further enhance the generation quality and efficiency. Additionally, the generated 3D avatars still lack hierarchical semantic structures and physical properties, which will be a direction worth exploring in future work.


\ifCLASSOPTIONcaptionsoff
  \newpage
\fi

{
\bibliographystyle{IEEEtran}
\bibliography{IEEEabrv,ref.bib}

\begin{thebibliography}{10}
\providecommand{\url}[1]{#1}
\csname url@samestyle\endcsname
\providecommand{\newblock}{\relax}
\providecommand{\bibinfo}[2]{#2}
\providecommand{\BIBentrySTDinterwordspacing}{\spaceskip=0pt\relax}
\providecommand{\BIBentryALTinterwordstretchfactor}{4}
\providecommand{\BIBentryALTinterwordspacing}{\spaceskip=\fontdimen2\font plus
\BIBentryALTinterwordstretchfactor\fontdimen3\font minus \fontdimen4\font\relax}
\providecommand{\BIBforeignlanguage}[2]{{%
\expandafter\ifx\csname l@#1\endcsname\relax
\typeout{** WARNING: IEEEtran.bst: No hyphenation pattern has been}%
\typeout{** loaded for the language `#1'. Using the pattern for}%
\typeout{** the default language instead.}%
\else
\language=\csname l@#1\endcsname
\fi
#2}}
\providecommand{\BIBdecl}{\relax}
\BIBdecl

\bibitem{nerf}
B.~Mildenhall, P.~P. Srinivasan, M.~Tancik, J.~T. Barron, R.~Ramamoorthi, and R.~Ng, ``{NeRF: Representing Scenes as Neural Radiance Fields for View Synthesis},'' \emph{Communications of the ACM}, vol.~65, no.~1, pp. 99--106, 2021.

\bibitem{neus}
P.~Wang, L.~Liu, Y.~Liu, C.~Theobalt, T.~Komura, and W.~Wang, ``{NeuS: Learning Neural Implicit Surfaces by Volume Rendering for Multi-view Reconstruction},'' \emph{Advances in Neural Information Processing Systems}, vol.~34, pp. 27\,171--27\,183, 2021.

\bibitem{dmtet}
T.~Shen, J.~Gao, K.~Yin, M.-Y. Liu, and S.~Fidler, ``Deep marching tetrahedra: a hybrid representation for high-resolution 3d shape synthesis,'' in \emph{Advances in Neural Information Processing Systems}, 2021.

\bibitem{3dgs}
B.~Kerbl, G.~Kopanas, T.~Leimk{\"u}hler, and G.~Drettakis, ``{3D Gaussian Splatting for Real-Time Radiance Field Rendering},'' \emph{ACM Transactions on Graphics}, vol.~42, no.~4, July 2023.

\bibitem{saito2019pifu}
S.~Saito, Z.~Huang, R.~Natsume, S.~Morishima, A.~Kanazawa, and H.~Li, ``{Pifu: Pixel-aligned implicit function for high-resolution clothed human digitization},'' in \emph{Proceedings of the IEEE/CVF International Conference on Computer Vision}, 2019, pp. 2304--2314.

\bibitem{xiu2022icon}
Y.~Xiu, J.~Yang, D.~Tzionas, and M.~J. Black, ``{Icon: Implicit clothed humans obtained from normals},'' in \emph{Proceedings of the IEEE/CVF Conference on Computer Vision and Pattern Recognition}.\hskip 1em plus 0.5em minus 0.4em\relax IEEE, 2022, pp. 13\,286--13\,296.

\bibitem{xiu2023econ}
Y.~Xiu, J.~Yang, X.~Cao, D.~Tzionas, and M.~J. Black, ``{Econ: Explicit clothed humans optimized via normal integration},'' in \emph{Proceedings of the IEEE/CVF Conference on Computer Vision and Pattern Recognition}, 2023, pp. 512--523.

\bibitem{weng2023personnerf}
C.-Y. Weng, P.~P. Srinivasan, B.~Curless, and I.~Kemelmacher-Shlizerman, ``{Personnerf: Personalized reconstruction from photo collections},'' in \emph{Proceedings of the IEEE/CVF Conference on Computer Vision and Pattern Recognition}, 2023, pp. 524--533.

\bibitem{wang2023complete}
J.~Wang, J.~S. Yoon, T.~Y. Wang, K.~K. Singh, and U.~Neumann, ``Complete 3d human reconstruction from a single incomplete image,'' in \emph{Proceedings of the IEEE/CVF Conference on Computer Vision and Pattern Recognition}, 2023, pp. 8748--8758.

\bibitem{weng2022humannerf}
C.-Y. Weng, B.~Curless, P.~P. Srinivasan, J.~T. Barron, and I.~Kemelmacher-Shlizerman, ``{Humannerf: Free-viewpoint rendering of moving people from monocular video},'' in \emph{Proceedings of the IEEE/CVF Conference on Computer Vision and Pattern Recognition}, 2022, pp. 16\,210--16\,220.

\bibitem{jiang2022neuman}
W.~Jiang, K.~M. Yi, G.~Samei, O.~Tuzel, and A.~Ranjan, ``{Neuman: Neural human radiance field from a single video},'' in \emph{Proceedings of the European conference on computer vision (ECCV)}.\hskip 1em plus 0.5em minus 0.4em\relax Springer, 2022, pp. 402--418.

\bibitem{yu2023monohuman}
Z.~Yu, W.~Cheng, X.~Liu, W.~Wu, and K.-Y. Lin, ``{MonoHuman: Animatable Human Neural Field from Monocular Video},'' \emph{arXiv preprint arXiv:2304.02001}, 2023.

\bibitem{qian20243dgsavatar}
Z.~Qian, S.~Wang, M.~Mihajlovic, A.~Geiger, and S.~Tang, ``{3DGS-Avatar: Animatable Avatars via Deformable 3D Gaussian Splatting},'' in \emph{Proceedings of the IEEE/CVF Conference on Computer Vision and Pattern Recognition}, 2024.

\bibitem{d3ga}
W.~Zielonka, T.~Bagautdinov, S.~Saito, M.~Zollhöfer, J.~Thies, and J.~Romero, ``{Drivable 3D Gaussian Avatars},'' \emph{arXiv preprint arXiv:2311.08581}, 2023.

\bibitem{zhao2022human}
F.~Zhao, Y.~Jiang, K.~Yao, J.~Zhang, L.~Wang, H.~Dai, Y.~Zhong, Y.~Zhang, M.~Wu, L.~Xu \emph{et~al.}, ``{Human Performance Modeling and Rendering via Neural Animated Mesh},'' \emph{ACM Transactions on Graphics (TOG)}, vol.~41, no.~6, pp. 1--17, 2022.

\bibitem{jiang2024uvgaussian}
Y.~Jiang, Q.~Liao, X.~Li, L.~Ma, Q.~Zhang, C.~Zhang, Z.~Lu, and Y.~Shan, ``{UV Gaussians: Joint Learning of Mesh Deformation and Gaussian Textures for Human Avatar Modeling},'' \emph{arXiv preprint arXiv:2403.11589}, 2024.

\bibitem{zheng2024physavatar}
Y.~Zheng, Q.~Zhao, G.~Yang, W.~Yifan, D.~Xiang, F.~Dubost, D.~Lagun, T.~Beeler, F.~Tombari, L.~Guibas \emph{et~al.}, ``{PhysAvatar: Learning the Physics of Dressed 3D Avatars from Visual Observations},'' \emph{arXiv preprint arXiv:2404.04421}, 2024.

\bibitem{dalle2}
A.~Ramesh, P.~Dhariwal, A.~Nichol, C.~Chu, and M.~Chen, ``{Hierarchical Text-Conditional Image Generation with CLIP Latents},'' \emph{arXiv preprint arXiv:2204.06125}, 2022.

\bibitem{latentdiffusion}
R.~Rombach, A.~Blattmann, D.~Lorenz, P.~Esser, and B.~Ommer, ``{High-Resolution Image Synthesis with Latent Diffusion Models},'' in \emph{Proceedings of the IEEE/CVF Conference on Computer Vision and Pattern Recognition}, 2022, pp. 10\,684--10\,695.

\bibitem{dreamfusion}
B.~Poole, A.~Jain, J.~T. Barron, and B.~Mildenhall, ``{DreamFusion: Text-to-3D using 2D Diffusion},'' \emph{arXiv preprint arXiv:2209.14988}, 2022.

\bibitem{sjc}
H.~Wang, X.~Du, J.~Li, R.~A. Yeh, and G.~Shakhnarovich, ``{Score Jacobian Chaining: Lifting Pretrained 2D Diffusion Models for 3D Generation},'' \emph{arXiv preprint arXiv:2212.00774}, 2022.

\bibitem{avatarclip}
F.~Hong, M.~Zhang, L.~Pan, Z.~Cai, L.~Yang, and Z.~Liu, ``{AvatarCLIP: Zero-Shot Text-Driven Generation and Animation of 3D Avatars},'' \emph{ACM Transactions on Graphics (TOG)}, vol.~41, no.~4, pp. 1--19, 2022.

\bibitem{jiang2023avatarcraft}
R.~Jiang, C.~Wang, J.~Zhang, M.~Chai, M.~He, D.~Chen, and J.~Liao, ``{AvatarCraft: Transforming Text into Neural Human Avatars with Parameterized Shape and Pose Control},'' \emph{arXiv preprint arXiv:2303.17606}, 2023.

\bibitem{liao2024tada}
T.~Liao, H.~Yi, Y.~Xiu, J.~Tang, Y.~Huang, J.~Thies, and M.~J. Black, ``{TADA! Text to Animatable Digital Avatars},'' in \emph{International Conference on 3D Vision (3DV)}, 2024.

\bibitem{kolotouros2024dreamhuman}
N.~Kolotouros, T.~Alldieck, A.~Zanfir, E.~Bazavan, M.~Fieraru, and C.~Sminchisescu, ``{DreamHuman: Animatable 3D Avatars from Text},'' \emph{Advances in Neural Information Processing Systems}, vol.~36, 2024.

\bibitem{yuan2024gavatar}
Y.~Yuan, X.~Li, Y.~Huang, S.~De~Mello, K.~Nagano, J.~Kautz, and U.~Iqbal, ``{GAvatar: Animatable 3D Gaussian Avatars with Implicit Mesh Learning},'' in \emph{Proceedings of the IEEE Conference on Computer Vision and Pattern Recognition}, 2024.

\bibitem{liu2024humangaussian}
X.~Liu, X.~Zhan, J.~Tang, Y.~Shan, G.~Zeng, D.~Lin, X.~Liu, and Z.~Liu, ``{HumanGaussian: Text-Driven 3D Human Generation with Gaussian Splatting},'' in \emph{Proceedings of the IEEE/CVF Conference on Computer Vision and Pattern Recognition}, 2024, pp. 6646--6657.

\bibitem{huang2023dreamwaltz}
Y.~Huang, J.~Wang, A.~Zeng, H.~Cao, X.~Qi, Y.~Shi, Z.-J. Zha, and L.~Zhang, ``{DreamWaltz: Make a Scene with Complex 3D Animatable Avatars},'' in \emph{Advances in Neural Information Processing Systems}, 2023.

\bibitem{controlnet}
L.~Zhang and M.~Agrawala, ``{Adding Conditional Control to Text-to-Image Diffusion Models},'' in \emph{Proceedings of the IEEE/CVF International Conference on Computer Vision}, 2023.

\bibitem{humansd}
X.~Ju, A.~Zeng, C.~Zhao, J.~Wang, L.~Zhang, and Q.~Xu, ``{HumanSD: A Native Skeleton-Guided Diffusion Model for Human Image Generation},'' in \emph{Proceedings of the IEEE/CVF International Conference on Computer Vision}, 2023.

\bibitem{smpl}
M.~Loper, N.~Mahmood, J.~Romero, G.~Pons-Moll, and M.~J. Black, ``{SMPL: a skinned multi-person linear mode},'' \emph{ACM transactions on graphics (TOG)}, vol.~34, no.~6, pp. 1--16, 2015.

\bibitem{smplx}
G.~Pavlakos, V.~Choutas, N.~Ghorbani, T.~Bolkart, A.~A. Osman, D.~Tzionas, and M.~J. Black, ``Expressive body capture: 3d hands, face, and body from a single image,'' in \emph{Proceedings of the IEEE/CVF Conference on Computer Vision and Pattern Recognition}, 2019, pp. 10\,975--10\,985.

\bibitem{instant-ngp}
T.~M{\"u}ller, A.~Evans, C.~Schied, and A.~Keller, ``{Instant Neural Graphics Primitives with a Multiresolution Hash Encoding},'' \emph{ACM Transactions on Graphics (ToG)}, vol.~41, no.~4, pp. 1--15, 2022.

\bibitem{glide}
A.~Nichol, P.~Dhariwal, A.~Ramesh, P.~Shyam, P.~Mishkin, B.~McGrew, I.~Sutskever, and M.~Chen, ``{GLIDE: Towards Photorealistic Image Generation and Editing with Text-Guided Diffusion Models},'' \emph{arXiv preprint arXiv:2112.10741}, 2021.

\bibitem{imagen}
C.~Saharia, W.~Chan, S.~Saxena, L.~Li, J.~Whang, E.~Denton, S.~K.~S. Ghasemipour, B.~K. Ayan, S.~S. Mahdavi, R.~G. Lopes \emph{et~al.}, ``{Photorealistic Text-to-Image Diffusion Models with Deep Language Understanding},'' \emph{arXiv preprint arXiv:2205.11487}, 2022.

\bibitem{beatsgan}
P.~Dhariwal and A.~Nichol, ``{Diffusion Models Beat GANs on Image Synthesis},'' \emph{{Advances in Neural Information Processing Systems}}, vol.~34, pp. 8780--8794, 2021.

\bibitem{ddim}
J.~Song, C.~Meng, and S.~Ermon, ``{Denoising Diffusion Implicit Models},'' in \emph{International Conference on Learning Representations}, 2021.

\bibitem{improved_ddpm}
A.~Q. Nichol and P.~Dhariwal, ``{Improved Denoising Diffusion Probabilistic Models},'' in \emph{International Conference on Machine Learning}.\hskip 1em plus 0.5em minus 0.4em\relax PMLR, 2021, pp. 8162--8171.

\bibitem{laion5b}
C.~Schuhmann, R.~Beaumont, R.~Vencu, C.~Gordon, R.~Wightman, M.~Cherti, T.~Coombes, A.~Katta, C.~Mullis, M.~Wortsman \emph{et~al.}, ``{LAION-5B: An open large-scale dataset for training next generation image-text models},'' \emph{arXiv preprint arXiv:2210.08402}, 2022.

\bibitem{cc}
P.~Sharma, N.~Ding, S.~Goodman, and R.~Soricut, ``{Conceptual Captions: A Cleaned, Hypernymed, Image Alt-text Dataset For Automatic Image Captioning},'' in \emph{Proceedings of the 56th Annual Meeting of the Association for Computational Linguistics (Volume 1: Long Papers)}, 2018, pp. 2556--2565.

\bibitem{cc12}
S.~Changpinyo, P.~Sharma, N.~Ding, and R.~Soricut, ``{Conceptual 12M: Pushing Web-Scale Image-Text Pre-Training To Recognize Long-Tail Visual Concepts},'' in \emph{Proceedings of the IEEE/CVF Conference on Computer Vision and Pattern Recognition}, 2021, pp. 3558--3568.

\bibitem{composer}
L.~Huang, D.~Chen, Y.~Liu, Y.~Shen, D.~Zhao, and J.~Zhou, ``{Composer: Creative and controllable image synthesis with composable conditions},'' in \emph{International Conference on Machine Learning}, 2023.

\bibitem{xiao2024ccm}
J.~Xiao, K.~Zhu, H.~Zhang, Z.~Liu, Y.~Shen, Z.~Yang, R.~Feng, Y.~Liu, X.~Fu, and Z.-J. Zha, ``{CCM: Real-Time Controllable Visual Content Creation Using Text-to-Image Consistency Models},'' in \emph{International Conference on Machine Learning}, 2024.

\bibitem{dit}
W.~Peebles and S.~Xie, ``{Scalable Diffusion Models with Transformers},'' in \emph{Proceedings of the IEEE/CVF International Conference on Computer Vision}, 2023, pp. 4195--4205.

\bibitem{sdxl}
D.~Podell, Z.~English, K.~Lacey, A.~Blattmann, T.~Dockhorn, J.~M{\"u}ller, J.~Penna, and R.~Rombach, ``{SDXL: Improving Latent Diffusion Models for High-Resolution Image Synthesis},'' \emph{arXiv preprint arXiv:2307.01952}, 2023.

\bibitem{sd3}
P.~Esser, S.~Kulal, A.~Blattmann, R.~Entezari, J.~M{\"u}ller, H.~Saini, Y.~Levi, D.~Lorenz, A.~Sauer, F.~Boesel \emph{et~al.}, ``{Scaling Rectified Flow Transformers for High-Resolution Image Synthesis},'' in \emph{International Conference on Machine Learning}, 2024.

\bibitem{hyperhuman}
X.~Liu, J.~Ren, A.~Siarohin, I.~Skorokhodov, Y.~Li, D.~Lin, X.~Liu, Z.~Liu, and S.~Tulyakov, ``{HyperHuman: Hyper-Realistic Human Generation with Latent Structural Diffusion},'' in \emph{International Conference on Learning Representations}, 2024.

\bibitem{objaverse}
M.~Deitke, D.~Schwenk, J.~Salvador, L.~Weihs, O.~Michel, E.~VanderBilt, L.~Schmidt, K.~Ehsani, A.~Kembhavi, and A.~Farhadi, ``{Objaverse: A Universe of Annotated 3D Objects},'' in \emph{Proceedings of the IEEE/CVF Conference on Computer Vision and Pattern Recognition}, 2023, pp. 13\,142--13\,153.

\bibitem{dreamfields}
A.~Jain, B.~Mildenhall, J.~T. Barron, P.~Abbeel, and B.~Poole, ``{Zero-Shot Text-Guided Object Generation With Dream Fields},'' in \emph{Proceedings of the IEEE/CVF Conference on Computer Vision and Pattern Recognition}, 2022, pp. 867--876.

\bibitem{clipmesh}
N.~Mohammad~Khalid, T.~Xie, E.~Belilovsky, and T.~Popa, ``{CLIP-Mesh: Generating textured meshes from text using pretrained image-text models},'' in \emph{SIGGRAPH Asia 2022 Conference Papers}, 2022, pp. 1--8.

\bibitem{clip}
A.~Radford, J.~W. Kim, C.~Hallacy, A.~Ramesh, G.~Goh, S.~Agarwal, G.~Sastry, A.~Askell, P.~Mishkin, J.~Clark \emph{et~al.}, ``{Learning Transferable Visual Models From Natural Language Supervision},'' in \emph{International Conference on Machine Learning}.\hskip 1em plus 0.5em minus 0.4em\relax PMLR, 2021, pp. 8748--8763.

\bibitem{magic3d}
C.-H. Lin, J.~Gao, L.~Tang, T.~Takikawa, X.~Zeng, X.~Huang, K.~Kreis, S.~Fidler, M.-Y. Liu, and T.-Y. Lin, ``{Magic3D: High-Resolution Text-to-3D Content Creation},'' \emph{arXiv preprint arXiv:2211.10440}, 2022.

\bibitem{fantasia3d}
R.~Chen, Y.~Chen, N.~Jiao, and K.~Jia, ``{Fantasia3D: Disentangling Geometry and Appearance for High-quality Text-to-3D Content Creation},'' \emph{arXiv preprint arXiv:2303.13873}, 2023.

\bibitem{dreamgaussian}
J.~Tang, J.~Ren, H.~Zhou, Z.~Liu, and G.~Zeng, ``{DreamGaussian: Generative Gaussian Splatting for Efficient 3D Content Creation},'' in \emph{International Conference on Learning Representations}, 2024.

\bibitem{dreamtime}
Y.~Huang, J.~Wang, Y.~Shi, B.~Tang, X.~Qi, and L.~Zhang, ``{DreamTime: An Improved Optimization Strategy for Diffusion-Guided 3D Generation},'' in \emph{International Conference on Learning Representations}, 2024.

\bibitem{nfsd}
O.~Katzir, O.~Patashnik, D.~Cohen-Or, and D.~Lischinski, ``{Noise-free Score Distillation},'' in \emph{International Conference on Learning Representations}, 2024.

\bibitem{csd}
X.~Yu, Y.-C. Guo, Y.~Li, D.~Liang, S.-H. Zhang, and X.~QI, ``Text-to-3d with classifier score distillation,'' in \emph{International Conference on Learning Representations}, 2024.

\bibitem{liang2023luciddreamer}
Y.~Liang, X.~Yang, J.~Lin, H.~Li, X.~Xu, and Y.~Chen, ``Luciddreamer: Towards high-fidelity text-to-3d generation via interval score matching,'' \emph{arXiv preprint arXiv:2311.11284}, 2023.

\bibitem{hifa}
J.~Zhu, P.~Zhuang, and S.~Koyejo, ``{HiFA: High-fidelity Text-to-3D Generation with Advanced Diffusion Guidance},'' in \emph{International Conference on Learning Representations}, 2024.

\bibitem{prolificdreamer}
Z.~Wang, C.~Lu, Y.~Wang, F.~Bao, C.~Li, H.~Su, and J.~Zhu, ``{ProlificDreamer: High-Fidelity and Diverse Text-to-3D Generation with Variational Score Distillation},'' in \emph{Advances in Neural Information Processing Systems}, 2023.

\bibitem{cao2023dreamavatar}
Y.~Cao, Y.-P. Cao, K.~Han, Y.~Shan, and K.-Y.~K. Wong, ``{DreamAvatar: Text-and-Shape Guided 3D Human Avatar Generation via Diffusion Models},'' \emph{arXiv preprint arXiv:2304.00916}, 2023.

\bibitem{zhang2024avatarverse}
H.~Zhang, B.~Chen, H.~Yang, L.~Qu, X.~Wang, L.~Chen, C.~Long, F.~Zhu, D.~Du, and M.~Zheng, ``{AvatarVerse: High-quality \& Stable 3D Avatar Creation from Text and Pose},'' in \emph{Proceedings of the AAAI Conference on Artificial Intelligence}, vol.~38, no.~7, 2024, pp. 7124--7132.

\bibitem{densepose}
R.~A. G{\"u}ler, N.~Neverova, and I.~Kokkinos, ``{DensePose: Dense Human Pose Estimation in the Wild},'' in \emph{Proceedings of the IEEE/CVF Conference on Computer Vision and Pattern Recognition}, 2018, pp. 7297--7306.

\bibitem{huang2024humannorm}
X.~Huang, R.~Shao, Q.~Zhang, H.~Zhang, Y.~Feng, Y.~Liu, and Q.~Wang, ``{HumanNorm: Learning Normal Diffusion Model for High-quality and Realistic 3D Human Generation},'' in \emph{Proceedings of the IEEE Conference on Computer Vision and Pattern Recognition}, 2024.

\bibitem{imghum}
T.~Alldieck, H.~Xu, and C.~Sminchisescu, ``{imGHUM: Implicit Generative Models of 3D Human Shape and Articulated Pose},'' in \emph{Proceedings of the IEEE/CVF International Conference on Computer Vision}, 2021, pp. 5461--5470.

\bibitem{yang2024deformable}
Z.~Yang, X.~Gao, W.~Zhou, S.~Jiao, Y.~Zhang, and X.~Jin, ``{Deformable 3D Gaussians for High-Fidelity Monocular Dynamic Scene Reconstruction},'' in \emph{Proceedings of the IEEE/CVF Conference on Computer Vision and Pattern Recognition}, 2024, pp. 20\,331--20\,341.

\bibitem{hu2024gaussianavatar}
L.~Hu, H.~Zhang, Y.~Zhang, B.~Zhou, B.~Liu, S.~Zhang, and L.~Nie, ``{GaussianAvatar: Towards Realistic Human Avatar Modeling from a Single Video via Animatable 3D Gaussians},'' in \emph{IEEE/CVF Conference on Computer Vision and Pattern Recognition}, 2024.

\bibitem{moon2024expressive}
G.~Moon, T.~Shiratori, and S.~Saito, ``Expressive whole-body 3d gaussian avatar,'' \emph{arXiv preprint arXiv:2407.21686}, 2024.

\bibitem{ddpm}
J.~Ho, A.~Jain, and P.~Abbeel, ``{Denoising Diffusion Probabilistic Models},'' \emph{Advances in Neural Information Processing Systems}, vol.~33, pp. 6840--6851, 2020.

\bibitem{stable-dreamfusion}
J.~Tang, ``Stable-dreamfusion: Text-to-3d with stable-diffusion,'' 2022, https://github.com/ashawkey/stable-dreamfusion.

\bibitem{latentnerf}
G.~Metzer, E.~Richardson, O.~Patashnik, R.~Giryes, and D.~Cohen-Or, ``{Latent-NeRF for Shape-Guided Generation of 3D Shapes and Textures},'' \emph{arXiv preprint arXiv:2211.07600}, 2022.

\bibitem{zeng2022deciwatch}
A.~Zeng, X.~Ju, L.~Yang, R.~Gao, X.~Zhu, B.~Dai, and Q.~Xu, ``{DeciWatch: A Simple Baseline for 10$\times$ Efficient 2D and 3D Pose Estimation},'' in \emph{Proceedings of the European conference on computer vision (ECCV)}.\hskip 1em plus 0.5em minus 0.4em\relax Springer, 2022, pp. 607--624.

\bibitem{mahmood2019amass}
N.~Mahmood, N.~Ghorbani, N.~F. Troje, G.~Pons-Moll, and M.~J. Black, ``{AMASS: Archive of motion capture as surface shapes},'' in \emph{Proceedings of the IEEE/CVF International Conference on Computer Vision}, 2019, pp. 5442--5451.

\bibitem{mohr2003building}
A.~Mohr and M.~Gleicher, ``{Building efficient, accurate character skins from examples},'' \emph{ACM Transactions on Graphics (TOG)}, vol.~22, no.~3, pp. 562--568, 2003.

\bibitem{pantazopoulos2002occlusion}
I.~Pantazopoulos and S.~Tzafestas, ``{Occlusion Culling Algorithms: A Comprehensive Survey},'' \emph{Journal of Intelligent and Robotic Systems}, vol.~35, pp. 123--156, 2002.

\bibitem{guedon2024sugar}
A.~Gu{\'e}don and V.~Lepetit, ``{SuGaR: Surface-Aligned Gaussian Splatting for Efficient 3D Mesh Reconstruction and High-Quality Mesh Rendering},'' in \emph{Proceedings of the IEEE/CVF Conference on Computer Vision and Pattern Recognition}, 2024, pp. 5354--5363.

\bibitem{waczynska2024games}
J.~Waczy{\'n}ska, P.~Borycki, S.~Tadeja, J.~Tabor, and P.~Spurek, ``{GaMeS: Mesh-Based Adapting and Modification of Gaussian Splatting},'' \emph{arXiv preprint arXiv:2402.01459}, 2024.

\bibitem{3dpw}
T.~Von~Marcard, R.~Henschel, M.~J. Black, B.~Rosenhahn, and G.~Pons-Moll, ``{Recovering Accurate 3D Human Pose in The Wild Using IMUs and a Moving Camera},'' in \emph{Proceedings of the European conference on computer vision (ECCV)}, 2018, pp. 601--617.

\bibitem{aist}
R.~Li, S.~Yang, D.~A. Ross, and A.~Kanazawa, ``{Learn to Dance with AIST++: Music Conditioned 3D Dance Generation},'' 2021.

\bibitem{motionx}
J.~Lin, A.~Zeng, S.~Lu, Y.~Cai, R.~Zhang, H.~Wang, and L.~Zhang, ``{Motion-X: A Large-scale 3D Expressive Whole-body Human Motion Dataset},'' in \emph{Advances in Neural Information Processing Systems}, 2023.

\bibitem{talkshow}
H.~Yi, H.~Liang, Y.~Liu, Q.~Cao, Y.~Wen, T.~Bolkart, D.~Tao, and M.~J. Black, ``{Generating Holistic 3D Human Motion from Speech},'' in \emph{Proceedings of the IEEE/CVF Conference on Computer Vision and Pattern Recognition}, 2023.

\bibitem{mip-nerf360}
J.~T. Barron, B.~Mildenhall, D.~Verbin, P.~P. Srinivasan, and P.~Hedman, ``{Mip-NeRF 360: Unbounded Anti-Aliased Neural Radiance Fields},'' in \emph{Proceedings of the IEEE/CVF conference on computer vision and pattern recognition}, 2022, pp. 5470--5479.

\end{thebibliography}
}

 




\begin{IEEEbiography}[{\includegraphics[width=1in,height=1.25in,clip,keepaspectratio]{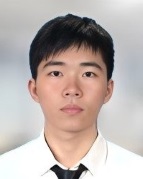}}] 
{Yukun Huang} is a Post-doctoral Research Fellow at the HKU Musketeers Foundation Institute of Data Science (HKU IDS). Previously, he obtained his PhD degree from the University of Science and Technology of China (USTC) and did his undergraduate studies at the South China University of Technology. His research interests broadly lie in the computer vision and machine learning. In particular, he is interested in 3D synthesis, virtual human, generative model, and person re-identification.
\end{IEEEbiography}

\begin{IEEEbiography}[{\includegraphics[width=1in,height=1.25in,clip,keepaspectratio]{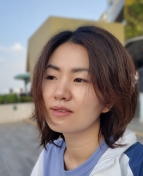}}] 
{Jianan Wang} received the MSc degree from the University of Oxford and currently serves as the chief researcher in AI cognition at Astribot. She has previously worked with DeepMind and the International Digital Economy Academy (IDEA). Her research interests and publications span computer vision and machine learning theory, with a recent focus on generative AI and robotics.
\end{IEEEbiography}

\begin{IEEEbiography}[{\includegraphics[width=1in,height=1.25in,clip,keepaspectratio]{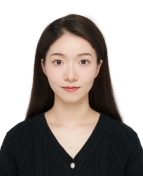}}] 
{Ailing Zeng} (Member, IEEE) is a senior researcher at Tencent AI Lab. Previously, she obtained her PhD degree from the Department of Computer Science and Engineering, the Chinese University of Hong Kong. Her research targets to build multi-modal human-like intelligent agents on scalable big data, especially for Large Motion Models to capture, understand, interact, and generate the motion of humans, animals, and the world. She has published over thirty top-tier conference papers at CVPR, NeurIPS, etc.
\end{IEEEbiography}

\begin{IEEEbiography}[{\includegraphics[width=1in,height=1.25in,clip,keepaspectratio]{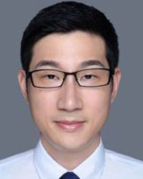}}] 
{Zheng-Jun Zha} (Member, IEEE) received the BE and PhD degrees from the University of Science and Technology of China, Hefei, China, in 2004 and 2009, respectively. He is currently a full professor with the School of Information Science and Technology, University of Science and Technology of China, and the executive director with the National Engineering Laboratory for Brain-Inspired Intelligence Technology and Application (NEL-BITA). He has authored or coauthored more than 200 papers in his research field with a series of publications on top journals and conferences, which include multimedia analysis and understanding, computer vision, pattern recognition, and brain-inspired intelligence. He was a recipient of multiple paper awards from prestigious conferences, including the Best Paper/Student Paper Award in Association for Computing Machinery (ACM) Multimedia and AAAI Distinguished Paper. He serves/served as an associated editor for IEEE Transactions on Multimedia, IEEE Transactions on Circuits and Systems for Video Technology, etc.
\end{IEEEbiography}

\begin{IEEEbiography}[{\includegraphics[width=1in,height=1.25in,clip,keepaspectratio]{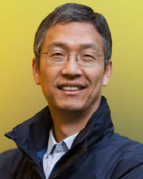}}] 
{Lei Zhang} (Fellow, IEEE) received the PhD degree in computer science from Tsinghua University, Beijing, China, in 2001. He is currently the chief scientist of computer vision and robotics with International Digital Economy Academy (IDEA) and an adjunct professor with the Hong Kong University of Science and Technology, Guangzhou, China. Prior to his current post, he was a principal researcher and research manager with Microsoft. He has authored or coauthored more than 150 techinical papers, and holds more than 60 U.S. patents in his research field, which include computer vision and machine learning, with particular focus on generic visual recognition at large scale. He was a editorial board member for IEEE Transactions on Multimedia, IEEE Transactions on Circuits and Systems for Video Technology, and Multimedia System Journal and as the area chair of many top conferences.
\end{IEEEbiography}

\begin{IEEEbiography}[{\includegraphics[width=1in,height=1.25in,clip,keepaspectratio]{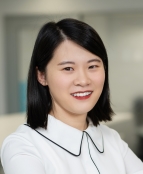}}] 
{Xihui Liu} (Member, IEEE) is an assistant professor at Department of Electrical and Electronic Engineering and Institute of Data Science, The University of Hong Kong. Before joining HKU, she was a postdoctoral researcher at University of California, Berkeley. She received the Bachelor's degree from Tsinghua University and PhD degree from The Chinese University of Hong Kong. Her research interests include computer vision, deep learning, generative models, and multimodal AI. She was awarded Adobe Research Fellowship 2020, EECS Rising Stars 2021, and WAIC Rising Star Award 2022. She serves as area chairs for CVPR 2024, ACM MM 2024, and ICLR 2025.
\end{IEEEbiography}

\end{document}